%% file: example.tex
\documentclass{article}
\usepackage{amssymb}
\usepackage{multirow}
\usepackage{booktabs} 
\usepackage{amsmath}  
\usepackage{graphicx}
\usepackage{capt-of}
\usepackage{algorithm}
\usepackage{algpseudocode}
\usepackage{etoc}
\usepackage{enumitem}
\usepackage{makecell}
\usepackage{wrapfig}
\usepackage{xspace}
\usepackage[table]{xcolor}
\usepackage{hyperref}
\usepackage{bm}
\usepackage{arydshln}
\usepackage[accsupp]{axessibility}
\usepackage[preprint]{corl_2026} 

\newcommand{\ours}[0]{\textsc{VAIC}\xspace}

\title{VAIC: Vision-Guided Humanoid Agile Object Interaction Control via Decoupled Commands}

\author{
Dongting Li$^{1,3*}$ \quad 
Qianyang Wu$^{3*}$ \quad 
Xingyu Chen$^{2}$ \quad 
Liang Li$^{3}$ \quad 
Yuhang Lin$^{3}$ 
\\
\textbf{Sikai Wu}$^{3}$ \quad 
\textbf{Guoyao Zhang}$^{3}$ \quad
\textbf{Mingliang Zhou}$^{3}$ \quad 
\textbf{Diyun Xiang}$^{3}$ \quad 
\textbf{Qiang Zhang}$^{2}$ 
\\
\textbf{Renjing Xu}$^{2}$ \quad 
\textbf{Jianzhu Ma}$^{1\dag}$  
\\
$^{1}$ Tsinghua University \quad 
$^{2}$ HKUST(Guangzhou) \quad 
$^{3}$ Xiaomi Robotics Lab
\\
$^*$Equal Contribution \quad $^{\dag}$ Corresponding author
\\
{\small \url{https://vaic-humanoid.github.io/}}
}

\begin{document}
\maketitle

\begin{figure}[h]
\centering
\vspace{-2em}
\includegraphics[width=\textwidth]{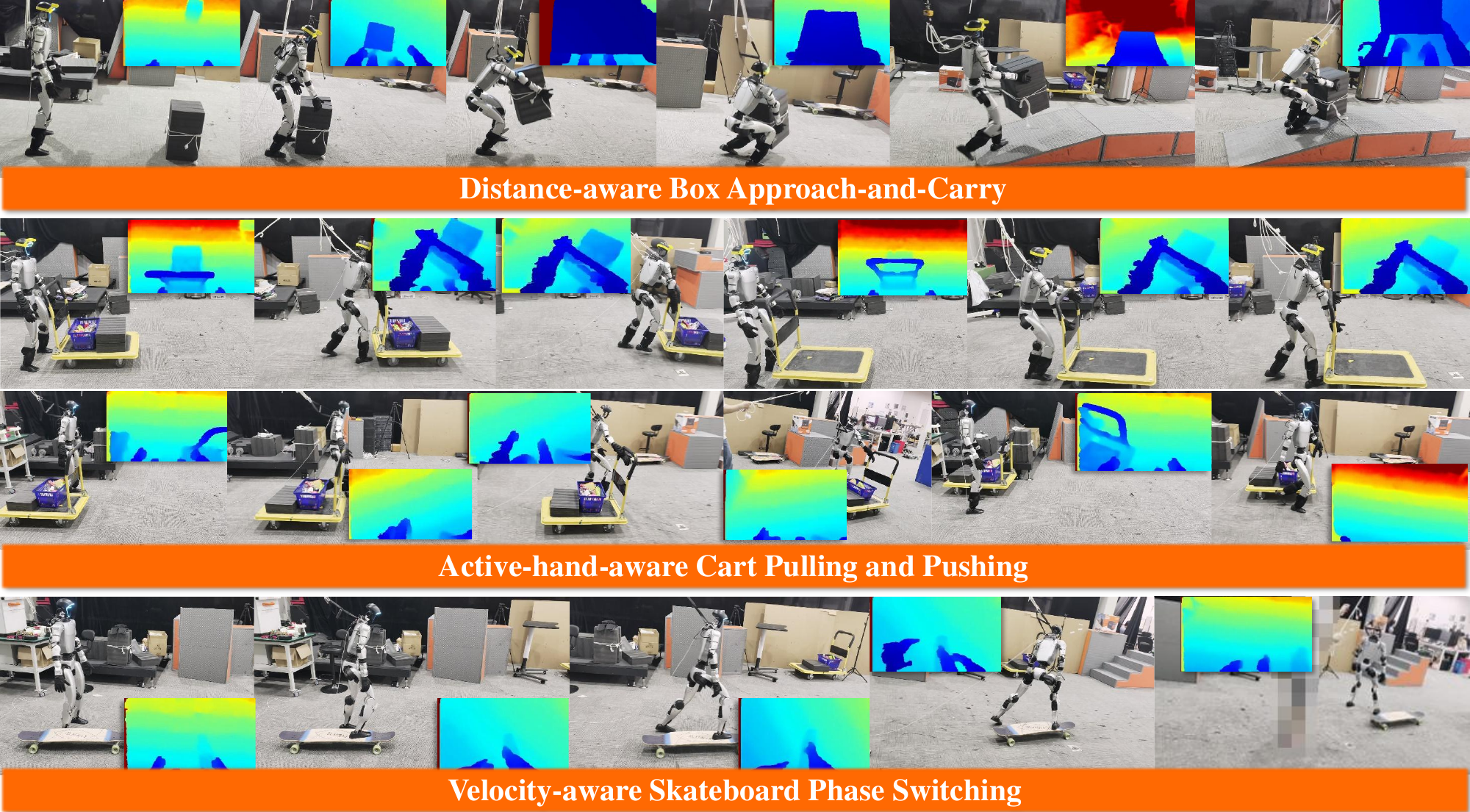}
\caption{We propose \ours, a unified vision-guided control framework that enables a humanoid robot to execute highly diverse and agile object interactions. By operating on onboard depth and decoupled user commands, VAIC successfully deploys dynamic tasks, such as distance-aware box carrying, underactuated cart interaction, and skateboarding.}
\label{fig:teaser}
\vspace{-1em}
\end{figure}

\begin{abstract}
    Humanoid robots hold immense potential for real-world assistance, yet agile interaction with objects in unstructured environments demands tightly coupled whole body coordination. Despite recent advancements, current controllers face a critical deployment gap. They rely heavily on dense reference trajectories and perfect state observability, which inherently limits physical generalization. We present Vision Guided Agile Interaction Control (VAIC), a unified framework that bridges this gap by operating exclusively on onboard depth, historical proprioception, and a decoupled user command interface. VAIC employs a two-stage distillation paradigm. First, a privileged teacher policy masters diverse interaction skills using precise object kinematics and exact environmental states. Second, a deployable student policy distills these capabilities by replacing full body tracking with velocity targets across multiple axes and an interaction indicator for each frame. The student utilizes a recurrent object adaptation module to implicitly infer unobservable object dynamics from raw depth streams and proprioception. Evaluations and real-world deployments on the humanoid robot demonstrate that a single VAIC policy successfully executes highly diverse dynamic tasks. These tasks include box carrying, cart interaction, and skateboarding, consistently outperforming baselines and advancing autonomous humanoid deployment.
\end{abstract}


\input{sections/1_introduction}
\input{sections/2_related_work}

\input{sections/3_methodology}
\input{sections/4_experiments}
\input{sections/5_conclusion}

\input{sections/6_limitation}




\bibliography{example}  

\input{sections/appendix}

\end{document}

%% file: sections/1_introduction.tex
\section{Introduction}
\label{sec:introduction}

Humanoid robots are increasingly expected to operate in unstructured human environments, executing tasks such as carrying bulky objects over stairs, manipulating underactuated carts, and adapting to abrupt dynamic disturbances. Unlike pure locomotion, these tasks demand tightly coupled whole body coordination, where the dynamic state of the interacting object fundamentally alters the robot's balance and contact mechanics.

Recent advances in deep reinforcement learning and motion imitation have enabled increasingly capable humanoid-object interaction \cite{wang2025physhsi, weng2025hdmilearninginteractivehumanoid, zhao2025resmimicgeneralmotiontracking, shao2025visualmimic}. Prior methods, such as HAIC \cite{li2026haic}, demonstrate that explicitly modeling coupled dynamics can enable agile interactions by tracking rich human demonstrations. However, deploying these capabilities in the physical world exposes a critical deployment gap characterized by two fundamental limitations. First, existing controllers typically demand dense, joint-level reference trajectories at runtime\cite{liao2025beyondmimicmotiontrackingversatile, luo2025sonic,weng2025hdmilearninginteractivehumanoid, li2026haic}. This kinematic hand holding is highly impractical for real-world autonomy or teleoperation, where human operators can only provide high-level intent, such as a joystick velocity command \cite{peng2021amp,lin2026pro}. Second, these methods heavily rely on perfect state observability, such as the ground truth 6D pose of objects and precise terrain geometry \cite{weng2025hdmilearninginteractivehumanoid, yang2025omniretarget, zhao2025resmimicgeneralmotiontracking,wang2026humanx}. In physical deployment, the robot's downward view is frequently occluded by the objects it carries, and real time state estimation is plagued by severe sensor noise and blind spots.

We address both limitations with \ours (Vision Guided Agile Interaction Control), a unified learning framework that bridges the gap between oracle-level simulation tracking and physical, perceptive deployment. The core insight of \ours is to decouple high-level user intent from low-level interaction execution. Instead of forcing the deployable policy to track a dense reference motion, we condition it on a unified, intent-driven command interface comprising a per-frame velocity command for navigational intent and a per-frame interaction state specifying the manipulation phase. To overcome severe visual occlusion without losing critical interaction priors, \ours employs a two-stage distillation paradigm. A privileged teacher policy first masters diverse interactive skills using exact object kinematics and terrain geometry. Subsequently, a deployable student policy distills these capabilities by learning to implicitly infer the missing privileged states, relying solely on temporal depth features, proprioceptive history, and the user commands via a recurrent Object Adaptation module.

This unified formulation allows a single policy to generalize across radically different dynamic profiles without task specific retraining. We validate \ours across three highly challenging categories of real-world interaction: (1) \textbf{Box Carrying} across varied terrain (flat, slopes, and stairs) under severe visual occlusion; (2) \textbf{Cart Manipulation}, managing the non holonomic constraints and external forces of an underactuated payload; and (3) \textbf{Skateboarding}, maintaining balance under impulsive contact dynamics. Concretely, our primary contributions are three points:



\begin{itemize}[leftmargin=0pt, label={}, labelsep=2pt, itemindent=0pt, parsep=0pt]
    \item \textbf{1) We propose a Decoupled Command Interface for Interaction.}
          We introduce a deployment friendly control formulation that replaces dense reference motion requirements with a high level command interface: multi axis velocity targets for locomotion intent and a per frame interaction state for phase control. This elegantly separates ``where to go'' from ``how to interact,'' enabling intuitive teleoperation and autonomous navigation.

    \item \textbf{2) We develop an Occlusion Resistant Distillation method.}
          To effectively utilize vision and handle imperfect observations, we introduce a depth recurrent Object Adaptation module. This module successfully distills privileged geometric and dynamic states from the teacher, allowing the robot to implicitly reconstruct unobservable object dynamics and execute precise interactions even when its exteroceptive vision is severely occluded or plagued by sensor noise.

    \item \textbf{3) We provide Unified Hardware Validation.}
          We demonstrate that a single \textit{ours} policy can generalize across diverse terrain sequences and seamlessly adapt to out-of-distribution object attributes. Extensive hardware deployments on the humanoid robot validate the system's robustness across carrying, pushing, pulling, and skateboarding behaviors, providing strong empirical evidence for its physical adaptability to unmodeled dynamic variations.
\end{itemize}

%% file: sections/2_related_work.tex
\section{Related Work}
\label{sec:related_work}
\subsection{Humanoid Whole-Body Control}
Physics-based motion imitation has established the foundation for acquiring rich humanoid motor skills. DeepMimic and AMP demonstrated that reference-conditioned training and adversarial motion priors can produce natural and dynamic whole-body behaviors~\cite{peng2018deepmimic,peng2021amp}. Building on this foundation, recent general motion trackers and retargeting frameworks further improve robustness and deployability through universal tracking, morphology-aware retargeting, residual correction, and simulation-to-real alignment~\cite{chen2025gmt,yin2025unitrackerlearninguniversalwholebody,joao2025gmr,he2024hover,he2025asap,liao2025beyondmimicmotiontrackingversatile,zhao2025resmimicgeneralmotiontracking}. More recent efforts scale such policies toward behavior priors and foundation-style controllers, including Kimodo~\cite{rempe2026kimodo} for large-scale controllable motion generation and SONIC~\cite{luo2025sonic} for supersized motion tracking with larger models, datasets, and compute, alongside promptable or generalist whole-body control models~\cite{zeng2025behavior,yuan2025behavior,li2025bfm,jiang2025uniact,shao2025langwbc}. However, these methods mainly focus on free-body motion skills, whereas our setting requires whole-body control under physical coupling with interactive objects.

\subsection{Humanoid-Object Interaction}
Learning to interact with objects requires jointly reasoning about whole-body contact, balance, and object dynamics. Large-scale human-object datasets provide diverse demonstrations for data-driven skill acquisition~\cite{bhatnagar2022behave, zhang2023neuraldome,lu2025humoto,zhao2024m, xu2025interact}, while simulated Humanoid-Object Interaction frameworks leverage such data for motion imitation, generative synthesis, or task-tokenized priors~\cite{wang2023physhoi,xu2025intermimic,lin2025simgenhoiphysicallyrealisticwholebody,wu2024thor,pan2025tokenhsi}. For real robots, recent loco-manipulation systems demonstrate box carrying, force-adaptive manipulation, unified whole-body control, heterogeneous meta-control, and dynamics-aware agile interaction~\cite{dao2024sim,zhang2025falcon,sun2025ulc,wei2025hmc,li2026haic}. Athletic humanoid systems further extend object interaction to highly dynamic tasks such as tennis, badminton, soccer, and ping-pong~\cite{zhang2026learning,chen2026learning,kong2026learning,ren2026smash}, while video-based imitation reduces manual motion engineering by extracting interactive priors from human demonstrations~\cite{weng2025hdmilearninginteractivehumanoid,allshire2025videomimic,yang2025omniretarget}. In contrast to methods that assume object states are available from simulation, motion capture, or external tracking\cite{zhao2025resmimicgeneralmotiontracking,wang2025physhsi,wang2026humanx,wu2026sugar}, \ours targets interaction with onboard object perception.

\subsection{Perceptive Humanoid Control}
Depth cameras, stereo sensors, and LiDAR have been integrated into locomotion policies to traverse stairs, gaps, slopes, and stepping stones~\cite{agarwal2023legged,long2024learning,sun2025dpl,han2025prior,zhang2026rpl,sun2026now}, while perceptive parkour systems combine depth sensing with dynamic motion tracking or skill composition to negotiate challenging obstacles~\cite{zhuang2024humanoid,wu2026perceptive,zhuang2026deep,zhu2026ttt}. In these works, perception is mainly used for foothold selection or obstacle avoidance in a static environment. Vision has also been introduced into humanoid-object interaction tasks.  Some methods explore latent vision-language instructions, visual sim-to-real transfer, and end-to-end VLA architectures for loco-manipulation~\cite{xue2025leverb,he2025viral,jiang2025wholebodyvlaunifiedlatentvla}, while ego-vision contact planning uses forward-looking depth to anticipate contact geometry~\cite{liu2025ego}. VisualMimic, LessMimic, Pro-HOI, and ULTRA use depth observations, perceptive object representations, or root-state guidance to support humanoid-object interaction~\cite{shao2025visualmimic,lin2026lessmimic,lin2026pro,he2026ultra}. Different from these approaches, \ours effectively combines vision and proprioceptive history to continuously infer the latent dynamics of the interacted object, enabling adaptive whole-body interaction with user-friendly command.


%% file: sections/3_methodology.tex
\section{Methodology}
\label{sec:methodology}

\begin{figure}[t]
\centering
\includegraphics[width=\textwidth]{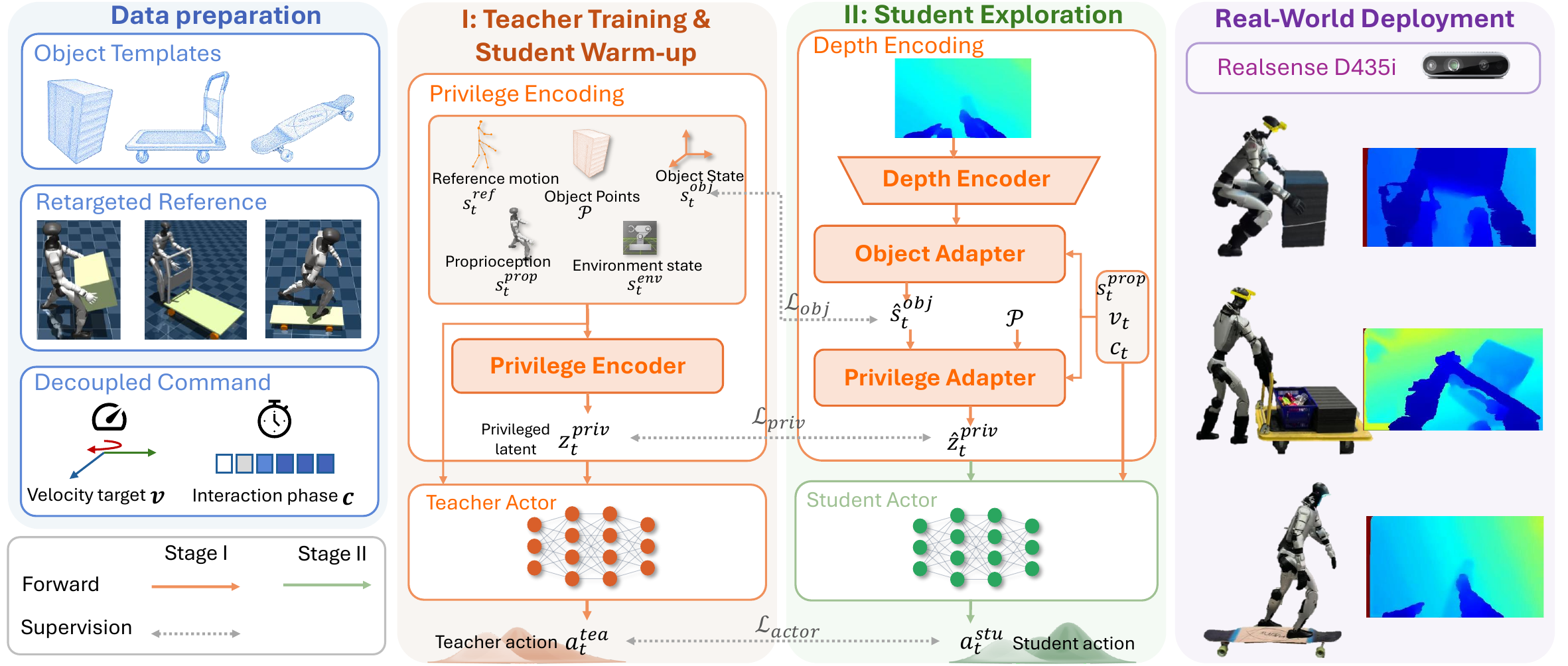}
\caption{Overview of \ours. The framework follows a two-stage distillation paradigm. A privileged teacher policy masters interaction skills using exact object states and reference motions. The deployable student policy strictly relies on onboard depth, proprioception, and a decoupled user command interface, distilling the teacher's capability via hierarchical alignment modules.} 
\label{fig:method_overview}
\vspace{-1em}
\end{figure}

Our goal is to train a unified whole-body controller capable of versatile object interactions in unstructured environments. As illustrated in Figure~\ref{fig:method_overview}, \ours bridges the gap between oracle simulation and real-world deployment through a specialized two-stage online distillation framework. By formalizing a decoupled user command interface, we explicitly eliminate the restrictive reliance on dense reference trajectories during physical deployment.

\subsection{Decoupled Command Interface}
We utilize a comprehensive dataset of human-object interactions retargeted to the humanoid robot (data preparation and geometric object template $\mathcal{P}$ extraction are detailed in Appendix~\ref{sec:appendix_data_prep}). 

Crucially, to facilitate intuitive teleoperation and autonomous deployment, we formalize a decoupled user command interface defined as $\mathbf{u}_t = (v_t, c_t)$. During the training phase, these commands are directly extracted from the retargeted reference motions to ensure temporal alignment with the demonstrated behaviors. This interface distills high-level task intent into two distinct components: (1) a multi-axis velocity target $v_t \in \mathbb{R}^3$, representing the navigational intent formatted as $[v_x, v_y, v_{\text{yaw}}]$, and (2) a per-frame binary indicator $c_t \in \{0, 1\}$ specifying the manipulation phase, where $0$ denotes navigating or approaching, and $1$ denotes active contact and interaction.

\subsection{Stage 1: Teacher Training and Student Warm-up}

In the first stage, we train a privileged teacher policy under an oracle simulation setting via reinforcement learning (PPO). As depicted in the Orange block of Figure~\ref{fig:method_overview}, a Privilege Encoder compresses the multimodal privileged context into a dense privileged latent $z_t^{priv}$:
\begin{equation}
    z_t^{priv} = f_{\text{enc\_priv}}(s_t^{ref}, s_t^{geo}, s_t^{prop}, s_t^{env}), \quad s_t^{geo} = \mathcal{T}(s_t^{obj}, \mathcal{P})
\end{equation}
Here, the transformed geometric feature $s_t^{geo}$ explicitly maps the nominal template $\mathcal{P}$ into the robot's local frame using the ground-truth object state $s_t^{obj}$ to provide exact spatial perception. The Teacher Actor then takes $z_t^{priv}$ and the proprioceptive state $s_t^{prop}$ to output the optimal joint action $a_t^{tea}$. Operating as an oracle motion tracker, the teacher can fully master the complex contact dynamics.

\textbf{Concurrent Distillation:} During the teacher's PPO exploration, we continuously utilize its rollout data to warm up the deployable student modules (Figure~\ref{fig:method_overview}, Green block). The Object Adapter and Privilege Adapter are trained via supervised learning ($\mathcal{L}_{obj} = \|\hat{s}_t^{obj} - s_t^{obj}\|^2$, $\mathcal{L}_{priv} = \|\hat{z}_t^{priv} - z_t^{priv}\|^2$) to implicitly infer the unobservable object dynamics and align the latent representations. To maximize computational efficiency during this stage, the visual input is zero-injected, meaning the depth tensor is structurally maintained but populated entirely with zeros. Simultaneously, the Student Actor undergoes behavior cloning (BC) via $\mathcal{L}_{actor} = \|a_t^{stu} - a_t^{tea}\|^2$. Crucially, this distillation features an asymmetric input design: while the teacher tracks the dense reference $s_t^{ref}$, the student actor is strictly conditioned on the corresponding decoupled commands $(v_t, c_t)$ extracted from the dataset, completely removing its dependence on the reference kinematics.

\subsection{Stage 2: Vision-Guided Student Exploration}

While Stage 1 establishes a robust prior, purely off-policy distillation suffers from compounding errors when the student encounters out-of-distribution states during real-world execution. In Stage 2, the dense reference motion and perfect $s_t^{obj}$ are completely removed from the deployable inputs. The Student Actor takes full control of environment exploration using PPO to optimize for task-completion and command-tracking rewards, while the preceding alignment modules continue to be trained via supervised learning using the exact states available during simulation rollouts.

During this active exploration phase, the hierarchical student modules operate as follows:

\noindent\textbf{Object Adapter:} A CNN-GRU Depth Encoder extracts temporal features from the noisy depth stream $\mathbf{d}_t$. The Object Adapter implicitly infers the dynamic object state:
\begin{equation}
    \hat{s}_t^{obj} = f_{\text{obj\_adapt}}(z_t^{depth}, s_t^{prop}, v_t, c_t), \quad z_t^{depth} = f_{\text{enc\_depth}}(\mathbf{d}_t)
\end{equation}
The binary indicator $c_t$ serves as a critical mode-switching prior, allowing the recurrent network to anticipate changes in coupled object dynamics and contact forces during interaction transitions.

\noindent\textbf{Privilege Adapter:} Using $\hat{s}_t^{obj}$ and the geometric template $\mathcal{P}$, this adapter projects spatial occupancy priors into the robot's frame to seamlessly reconstruct the policy latent:
\begin{equation}
    \hat{z}_t^{priv} = f_{\text{priv\_adapt}}(s_t^{prop}, \hat{s}_t^{geo}, v_t, c_t), \quad \hat{s}_t^{geo} = \mathcal{T}(\hat{s}_t^{obj}, \mathcal{P})
\end{equation}

\noindent\textbf{Online Adaptation:} Crucially, as the Student Actor actively explores and generates new trajectories, the Object Adapter and Privilege Adapter continue to be updated online using the student's own rollout data. This tight coupling prevents representation drift and strictly grounds the visual reconstruction in the student's actual physical interactions.

%% file: sections/4_experiments.tex
\section{Experiments}
\label{sec:experiments}

\subsection{Experimental Setup}

\subsubsection{Task Description}
We evaluate \ours on three challenging categories of whole-body interaction:

\noindent\textbf{Box Interaction.} The robot carries a bulky box across varying terrain profiles, including flat ground, slopes, stairs, and composite sequences. The box severely occludes the robot's downward view, requiring the policy to rely on depth-based implicit state reconstruction for terrain-aware locomotion.

\noindent\textbf{Cart Interaction.} The robot manipulates a wheeled cart via pulling and pushing motions. The cart's non-holonomic constraints and underactuated dynamics introduce time-varying external disturbances that actively affect the robot's balance.

\noindent\textbf{Skateboard Interaction.} The robot balances on and rides a skateboard, including high-speed gliding and dynamic dismount phases. This demands precise anticipation of the board's translational and rotational dynamics under impulsive forces.

\subsubsection{Implementation Details}
We conduct experiments on a humanoid robot equipped with an onboard Intel RealSense D435i camera. The policy is executed at 50 Hz, while the low-level joint controller runs at 200 Hz. Depth observations from the RealSense camera are captured at 60 Hz.

\noindent\textbf{Data Collection.} Human demonstrations were captured via an optical motion capture system tracking both the performer and the interactive objects. Raw kinematic data were retargeted to the humanoid robot and refined through a physics-based simulation pipeline to generate physically valid states with precise contact information.

\noindent\textbf{Training.} Both training stages were executed using PPO in Isaac Sim on a single NVIDIA RTX 4090 GPU. Stage 1 (Teacher) utilizes 4096 parallel environments; Stage 2 (Student Fine-tuning) utilizes 512 environments with real-time onboard camera rendering enabled. Stage 1 converges in approximately 5 hours. Stage 2 converges in approximately 15 hours.

\subsubsection{Evaluation Metrics}
All quantitative simulation comparisons are rigorously evaluated in MuJoCo. To provide a concise overview, metrics are averaged across all subtask sequences within each object category. We report the following three categories of metrics. (1) \textbf{Success Rate (SR):} An episode is successful if the robot maintains dynamic balance, executes the decoupled command $(v_t, c_t)$, and avoids losing control of the object. (2) \textbf{Root State Metrics:} These include tracking errors for the humanoid root in position ($E_\mathrm{rpe}$), orientation ($E_\mathrm{roe}$), linear velocity ($E_\mathrm{rve}$), angular velocity ($E_\mathrm{rave}$), and linear acceleration ($E_\mathrm{rae}$). (3) \textbf{Object State Metrics:} These comprise the corresponding tracking errors for the interactive object ($E_\mathrm{ope}$, $E_\mathrm{ooe}$, $E_\mathrm{ove}$, $E_\mathrm{oave}$, $E_\mathrm{oae}$).

\input{tables/box_sim_experiments}
\input{tables/cart_sim_experiments}
\input{tables/skateboard_sim_experiments}

\subsubsection{Baselines}
We compare \ours in simulation against four representative whole body control and interaction baselines. (1) \textbf{PPO} \cite{schulman2017proximalpolicyoptimizationalgorithms}: A standard policy trained directly with dense reference motions, lacking the occlusion resistant distillation. (2) \textbf{AMP} \cite{peng2021amp}: An adversarial motion prior framework designed for kinematic style matching. (3) \textbf{PhysHSI} \cite{wang2025physhsi} and (4) \textbf{VisualMimic} \cite{shao2025visualmimic}: Two state of the art interactive whole body controllers. The PPO and AMP baselines are retrained from scratch on our motion dataset using an identical reward structure to ensure a fair comparison.

\subsection{Quantitative Evaluation}

We present the quantitative simulation results across our three primary evaluation categories. Specifically, Table~\ref{tab:sim_box} details the performance metrics for the Box Interaction task across varied terrains. Table~\ref{tab:sim_cart} reports the results for the Cart Interaction task under underactuated dynamics. Table~\ref{tab:sim_skateboard} summarizes the evaluation on the Skateboard Interaction task involving high-speed impulsive contacts. By conditioning on the decoupled command interface rather than blindly tracking full-body reference kinematics, \ours systematically outperforms all baselines. Pure kinematic trackers like PPO and AMP suffer catastrophic failure rates under dynamic disturbances due to their inability to implicitly estimate unobservable object states. In contrast, \ours maintains superior balance and achieves the lowest tracking errors across almost all root and object metrics.

To further evaluate robustness against spatial initialization and object variations, we compare \ours against PhysHSI and VisualMimic specifically on the Box Carry task. As detailed in Table~\ref{tab:box_carry_comparison}, we evaluate across four configurations comprising combinations of two box sizes and two initial distances. While baseline interactive methods struggle with modified initializations, \ours consistently achieves superior success rates, demonstrating a implicit understanding of coupled dynamics.

\input{tables/success_rate_comparison}

\subsection{Generalization to Unseen Dynamics}

\begin{figure}[t]
\centering
\includegraphics[width=\textwidth]{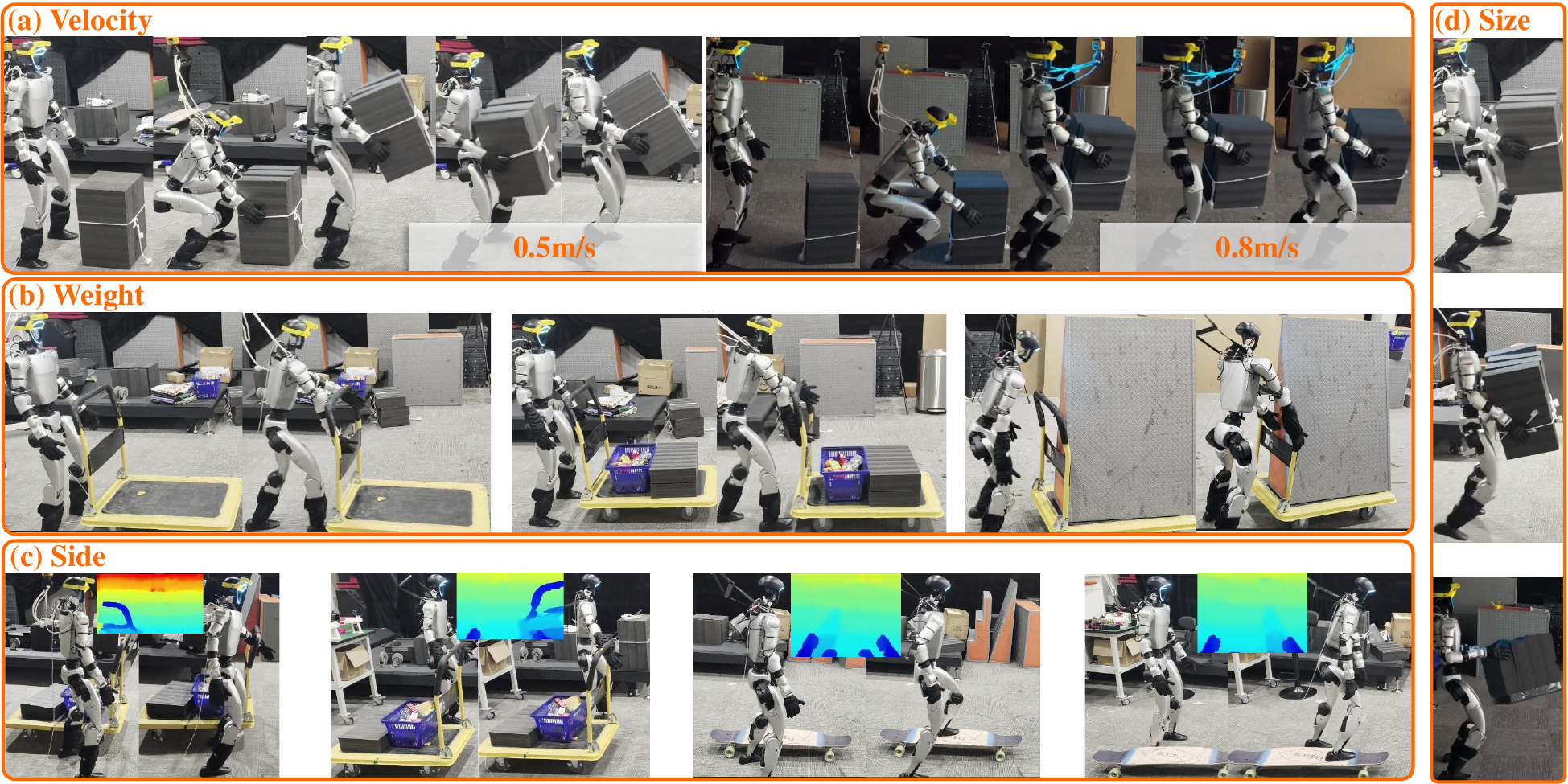}
\caption{Real-world hardware generalization of \ours to out-of-distribution object attributes and spatial conditions. The evaluations include (a) velocity command variations, (b) cart payload weight variations, (c) cart and skateboard position variations, and (d) box size variations. \ours robustly adapts its whole-body posture to maintain stable interactions during physical deployment.}
\label{fig:generalization}
\vspace{-1em}
\end{figure}

While the aforementioned quantitative tracking evaluations were conducted in the MuJoCo simulator, the ultimate validation of our framework lies in physical deployment. A key advantage of replacing explicit kinematic tracking with intent-driven distillation is the emergence of robust physical adaptability to unmodeled variations on real hardware. As demonstrated in Figure~\ref{fig:generalization}, the deployed \ours policy successfully handles out-of-distribution velocity commands, unmodeled payload weights, varying interaction sides, and different box sizes. The Object Adaptation module implicitly infers these altered physical relationships from history, allowing the humanoid robot to dynamically shift its center of gravity and adjust grasping postures in the real world.

\subsection{Mechanism Analysis}

\begin{figure}[t]
\centering
\includegraphics[width=\textwidth]{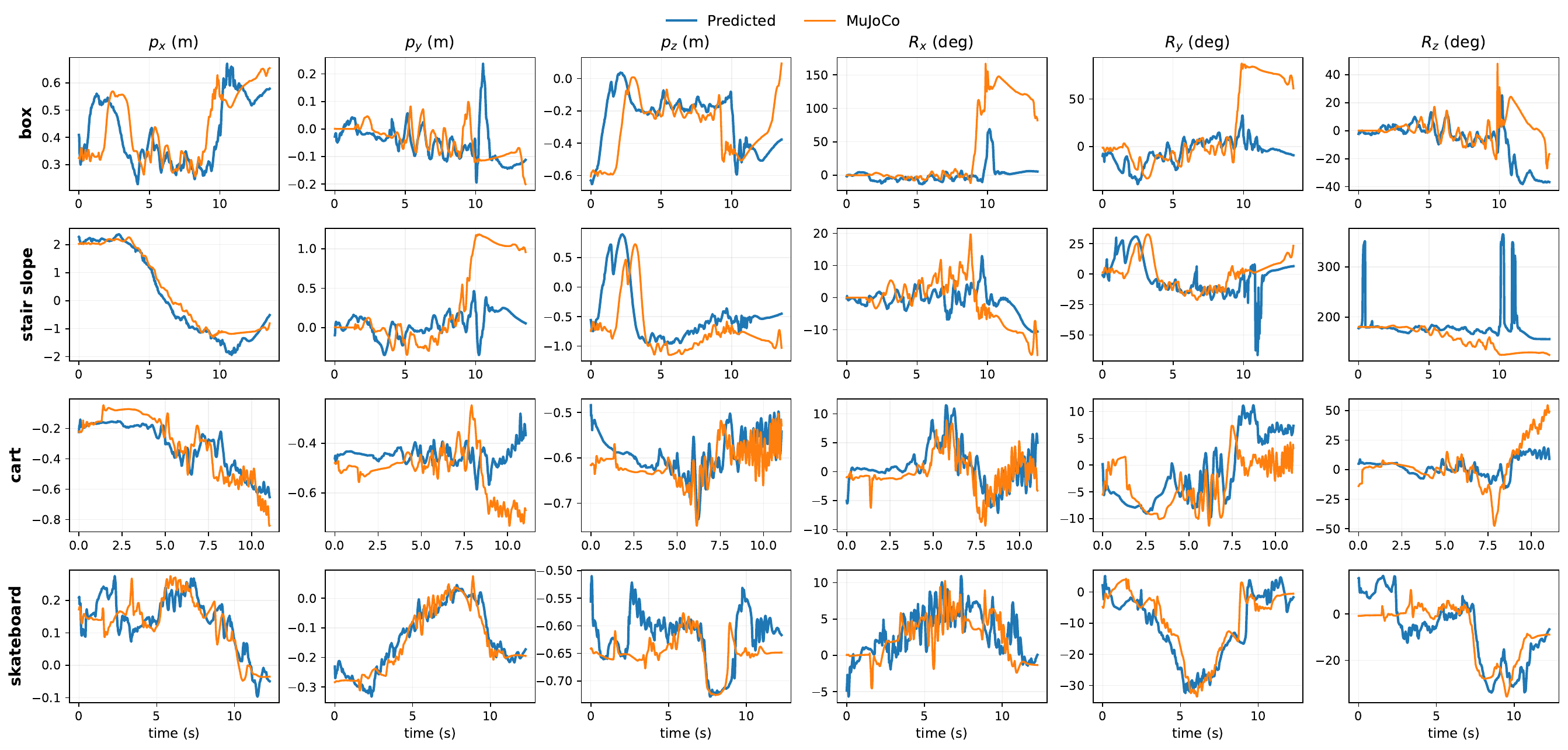}
\caption{Comparison of the predicted object state from the \ours adaptation module against MuJoCo ground truth. Conditioning the estimator on the binary interaction indicator $c_{\text{t}}$ prevents representation collapse during high-speed dynamic phases.}
\label{fig:state_estimation}
\vspace{-1em}
\end{figure}

We analyze the internal mechanism underlying this robustness by comparing the predicted object states $\hat{s}_t^{obj}$ from the Object Adaptation module of \ours against the MuJoCo ground truth. As demonstrated in Figure~\ref{fig:state_estimation}, the depth-conditioned estimator tightly tracks the object's position and orientation even during severe self-occlusion (e.g., box carrying over slopes and stairs) and dynamic dismounts (e.g., pulling a cart and skateboarding). Crucially, conditioning the estimator on the binary interaction phase $c_t$ significantly reduces temporal drift compared to purely proprioceptive alternatives.

%% file: tables/box_sim_experiments.tex
\begin{table*}[h]
\vspace{-0.5em}
\centering
\colorlet{lightorange}{orange!25}
\colorlet{lighterorange}{orange!12}
\caption{Quantitative evaluation on \textbf{Box Interaction} (averaged across flat, slope, and stair terrains). \ours leverages depth-based state prediction to effectively overcome severe downward visual occlusion caused by the carried box, achieving the highest success rate. Evaluation tracks both the humanoid root and the object state errors.}
\label{tab:sim_box}
\resizebox{\textwidth}{!}{
\begin{tabular}{lc ccccc ccccc}
\toprule
& & \multicolumn{5}{c}{\textbf{Root State Metrics ($\downarrow$)}} & \multicolumn{5}{c}{\textbf{Object State Metrics ($\downarrow$)}} \\
\cmidrule(lr){3-7} \cmidrule(lr){8-12}
\textbf{Method} & \textbf{SR} $\uparrow$ & \textbf{$E_\mathrm{rpe}$} & \textbf{$E_\mathrm{roe}$} & \textbf{$E_\mathrm{rve}$} & \textbf{$E_\mathrm{rave}$} & \textbf{$E_\mathrm{rae}$} & \textbf{$E_\mathrm{ope}$} & \textbf{$E_\mathrm{ooe}$} & \textbf{$E_\mathrm{ove}$} & \textbf{$E_\mathrm{oave}$} & \textbf{$E_\mathrm{oae}$} \\
\midrule
PPO   & 0.0\% & 1.83$^{\pm 0.40}$ & 0.84$^{\pm 0.07}$ & 0.51$^{\pm 0.08}$ & 0.43$^{\pm 0.05}$ & 0.55$^{\pm 0.06}$ & 0.94$^{\pm 0.40}$ & 0.89$^{\pm 0.35}$ & 0.27$^{\pm 0.07}$ & 0.40$^{\pm 0.08}$ & 0.43$^{\pm 0.11}$ \\
AMP   & 50.0\% & 0.78$^{\pm 0.22}$ & 0.61$^{\pm 0.30}$ & \textbf{0.30$^{\pm 0.07}$} & 0.42$^{\pm 0.11}$ & \textbf{0.42$^{\pm 0.08}$} & 1.01$^{\pm 0.46}$ & \textbf{0.65$^{\pm 0.50}$} & 0.27$^{\pm 0.10}$ & 0.25$^{\pm 0.12}$ & \textbf{0.27$^{\pm 0.09}$} \\
\textbf{\ours} & \cellcolor{lightorange}\textbf{66.7\%} & \cellcolor{lightorange}\textbf{0.59$^{\pm 0.22}$} & \cellcolor{lightorange}\textbf{0.51$^{\pm 0.14}$} & \cellcolor{lighterorange}0.33$^{\pm 0.05}$ & \cellcolor{lightorange}\textbf{0.41$^{\pm 0.05}$} & \cellcolor{lighterorange}0.53$^{\pm 0.06}$ & \cellcolor{lightorange}\textbf{0.62$^{\pm 0.09}$} & \cellcolor{lighterorange}0.69$^{\pm 0.38}$ & \cellcolor{lightorange}\textbf{0.21$^{\pm 0.02}$} & \cellcolor{lightorange}\textbf{0.25$^{\pm 0.09}$} & \cellcolor{lighterorange}0.28$^{\pm 0.03}$ \\
\bottomrule
\end{tabular}}
\vspace{-1em}
\end{table*}

%% file: tables/cart_sim_experiments.tex
\begin{table*}[h]
\centering
\colorlet{lightorange}{orange!25}
\colorlet{lighterorange}{orange!12}
\caption{Quantitative evaluation on \textbf{Cart Interaction} (averaged across pull and push tasks). Underactuated cart dynamics cause standard imitation baselines to frequently fail. \ours achieves superior success rates and minimal positional error, effectively anticipating the cart's physical feedback.}
\label{tab:sim_cart}
\resizebox{\textwidth}{!}{
\begin{tabular}{lc ccccc ccccc}
\toprule
& & \multicolumn{5}{c}{\textbf{Root State Metrics ($\downarrow$)}} & \multicolumn{5}{c}{\textbf{Object State Metrics ($\downarrow$)}} \\
\cmidrule(lr){3-7} \cmidrule(lr){8-12}
\textbf{Method} & \textbf{SR} $\uparrow$ & \textbf{$E_\mathrm{rpe}$} & \textbf{$E_\mathrm{roe}$} & \textbf{$E_\mathrm{rve}$} & \textbf{$E_\mathrm{rave}$} & \textbf{$E_\mathrm{rae}$} & \textbf{$E_\mathrm{ope}$} & \textbf{$E_\mathrm{ooe}$} & \textbf{$E_\mathrm{ove}$} & \textbf{$E_\mathrm{oave}$} & \textbf{$E_\mathrm{oae}$} \\
\midrule
PPO   & 41.7\% & 1.01$^{\pm 0.48}$ & 0.49$^{\pm 0.07}$ & 0.38$^{\pm 0.12}$ & 0.32$^{\pm 0.06}$ & 0.32$^{\pm 0.11}$ & 0.83$^{\pm 0.57}$ & 0.04$^{\pm 0.04}$ & 0.31$^{\pm 0.13}$ & 0.02$^{\pm 0.02}$ & \textbf{0.20$^{\pm 0.05}$} \\
AMP   & 58.3\% & 0.61$^{\pm 0.48}$ & 0.54$^{\pm 0.41}$ & 0.23$^{\pm 0.12}$ & \textbf{0.24$^{\pm 0.07}$} & \textbf{0.27$^{\pm 0.12}$} & 0.76$^{\pm 0.51}$ & \textbf{0.03$^{\pm 0.03}$} & \textbf{0.26$^{\pm 0.10}$} & \textbf{0.02$^{\pm 0.02}$} & 0.20$^{\pm 0.06}$ \\
\textbf{\ours} & \cellcolor{lightorange}\textbf{91.7\%} & \cellcolor{lightorange}\textbf{0.45$^{\pm 0.09}$} & \cellcolor{lightorange}\textbf{0.33$^{\pm 0.13}$} & \cellcolor{lightorange}\textbf{0.22$^{\pm 0.09}$} & \cellcolor{lighterorange}0.26$^{\pm 0.12}$ & \cellcolor{lighterorange}0.30$^{\pm 0.09}$ & \cellcolor{lightorange}\textbf{0.61$^{\pm 0.12}$} & \cellcolor{lighterorange}0.10$^{\pm 0.12}$ & \cellcolor{lighterorange}0.28$^{\pm 0.11}$ & \cellcolor{lighterorange}0.04$^{\pm 0.04}$ & \cellcolor{lighterorange}0.22$^{\pm 0.06}$ \\
\bottomrule
\end{tabular}}
\vspace{-1em}
\end{table*}

%% file: tables/skateboard_sim_experiments.tex
\begin{table*}[ht]
\centering
\colorlet{lightorange}{orange!25}
\colorlet{lighterorange}{orange!12}
\caption{Quantitative evaluation on \textbf{Skateboard Interaction}. Impulsive contact forces disrupt the balance of pure kinematic trackers, whereas \ours precisely tracks high-speed dynamics and achieves the highest success rate.}
\label{tab:sim_skateboard}
\resizebox{\textwidth}{!}{
\begin{tabular}{lc ccccc ccccc}
\toprule
& & \multicolumn{5}{c}{\textbf{Root State Metrics ($\downarrow$)}} & \multicolumn{5}{c}{\textbf{Object State Metrics ($\downarrow$)}} \\
\cmidrule(lr){3-7} \cmidrule(lr){8-12}
\textbf{Method} & \textbf{SR} $\uparrow$ & \textbf{$E_\mathrm{rpe}$} & \textbf{$E_\mathrm{roe}$} & \textbf{$E_\mathrm{rve}$} & \textbf{$E_\mathrm{rave}$} & \textbf{$E_\mathrm{rae}$} & \textbf{$E_\mathrm{ope}$} & \textbf{$E_\mathrm{ooe}$} & \textbf{$E_\mathrm{ove}$} & \textbf{$E_\mathrm{oave}$} & \textbf{$E_\mathrm{oae}$} \\
\midrule
PPO   & 0.0\% & 2.09$^{\pm 0.34}$ & 0.99$^{\pm 0.33}$ & 0.50$^{\pm 0.05}$ & 0.39$^{\pm 0.05}$ & 0.45$^{\pm 0.04}$ & 1.75$^{\pm 0.19}$ & 0.12$^{\pm 0.16}$ & 0.39$^{\pm 0.04}$ & 0.05$^{\pm 0.05}$ & \textbf{0.26$^{\pm 0.03}$} \\
AMP   & 12.5\% & 1.30$^{\pm 0.70}$ & 0.98$^{\pm 0.33}$ & 0.40$^{\pm 0.11}$ & 0.44$^{\pm 0.09}$ & 0.40$^{\pm 0.04}$ & 1.20$^{\pm 0.62}$ & 0.19$^{\pm 0.27}$ & 0.36$^{\pm 0.08}$ & 0.07$^{\pm 0.10}$ & 0.28$^{\pm 0.05}$ \\
\textbf{\ours} & \cellcolor{lightorange}\textbf{83.3\%} & \cellcolor{lightorange}\textbf{0.62$^{\pm 0.12}$} & \cellcolor{lightorange}\textbf{0.38$^{\pm 0.19}$} & \cellcolor{lightorange}\textbf{0.29$^{\pm 0.05}$} & \cellcolor{lightorange}\textbf{0.27$^{\pm 0.07}$} & \cellcolor{lightorange}\textbf{0.35$^{\pm 0.07}$} & \cellcolor{lightorange}\textbf{0.62$^{\pm 0.17}$} & \cellcolor{lightorange}\textbf{0.07$^{\pm 0.06}$} & \cellcolor{lightorange}\textbf{0.31$^{\pm 0.10}$} & \cellcolor{lightorange}\textbf{0.02$^{\pm 0.02}$} & \cellcolor{lighterorange}0.27$^{\pm 0.04}$ \\
\bottomrule
\end{tabular}}
\end{table*}

%% file: tables/success_rate_comparison.tex
\begin{table}[h]
\centering
\caption{Success rate comparison on the Box Carry task under varying initial conditions in MuJoCo. We evaluate physical robustness across combinations of two object sizes (Small, Large) and two initial interaction distances (Near, Far).}
\label{tab:box_carry_comparison}
\resizebox{0.5\textwidth}{!}{
\begin{tabular}{lcccc}
\toprule
Method & Small-Near & Small-Far & Large-Near & Large-Far \\
\midrule
PhysHSI & 9/10 & 9/10 & 8/10 & \textbf{9/10} \\
VisualMimic & - & - & \textbf{10/10} & 8/10 \\
\ours  & \textbf{10/10} & \textbf{10/10} & \textbf{10/10} & 7/10 \\
\bottomrule
\end{tabular}}
\vspace{-1em}
\end{table}

%% file: sections/5_conclusion.tex
\section{Conclusion}
\label{sec:conclusion}

In this paper, we presented \ours (Vision-Guided Agile Interaction Control), a unified framework for humanoid whole-body loco-manipulation in unstructured environments. By introducing a high-level decoupled command interface and an asymmetric two-stage distillation paradigm, \ours successfully eliminates the restrictive reliance on dense joint-level reference trajectories and perfect state observability during deployment. Facilitated by a recurrent Object Adaptation module, our policy effectively overcomes severe visual occlusion and implicitly predicts unmodeled coupled object dynamics. Extensive quantitative evaluations in simulation and real-world deployments on the humanoid robot across box carrying, cart manipulation, and skateboarding demonstrate the framework's superior balance, adaptability, and robust generalization, presenting a practical paradigm for autonomous humanoid deployment.

%% file: sections/6_limitation.tex
\section{Limitation}
\label{sec:limitation}

While \ours demonstrates robust generalization in physical deployments, several limitations remain. First, the depth-based Object Adaptation module is inherently susceptible to sensor noise and may fail when interacting with transparent or highly reflective objects. Second, although the recurrent architecture effectively infers unmodeled dynamics from proprioceptive history, extreme out-of-distribution physical variations (e.g., sudden and drastic mass shifts) can still induce representation collapse before the humanoid can adjust its balance. Finally, the decoupled command interface $(v_t, c_t)$ is highly efficient for navigational intent and gross loco-manipulation, but it lacks the expressive granularity required for dexterous, fine-grained in-hand manipulation. Future work will explore incorporating tactile feedback and richer semantic command spaces to address these constraints.

%% file: sections/appendix.tex
\clearpage
\setcounter{page}{1}
\vbox{%
  \hsize\textwidth
  \linewidth\hsize
  \vskip 0.1in
  \centering
  {\LARGE\bf VAIC: Vision-Guided Humanoid Agile Object Interaction Control via Decoupled Commands \par}
  \vskip 0.2in
  {\Large \bf Appendix\par}
  \vspace{0.3in minus 0.1in}
}

\vspace{2em}

\setcounter{table}{0}
\renewcommand{\thetable}{\Alph{table}}
\renewcommand*{\theHtable}{\thetable}
\setcounter{figure}{0}
\renewcommand{\thefigure}{\Alph{figure}}
\renewcommand*{\theHfigure}{\thefigure}
\setcounter{section}{0}
\renewcommand{\thesection}{\Alph{section}}
\renewcommand*{\theHsection}{\thesection}

\noindent In this appendix, we provide additional experimental setups and details:
\begin{enumerate}
    \item \textbf{Demo Video.} A demonstration video including the real-world experiment is provided, as described in Sec.~\ref{sec:demo}.
    \item \textbf{System Architecture and Parameters.} The observation and action spaces, alongside network architectures, are detailed in Sec.~\ref{sec:appendix_system}.
    \item \textbf{Experimental Details.} The reward formulation, training hyperparameters, real-world deployment setups, and mathematical definitions of our quantitative metrics are provided in Sec.~\ref{sec:appendix_exp_details}.
    \item \textbf{Data Preparation and Nominal Templates.} Details regarding motion retargeting, object template generation, and the extraction of unified decoupled commands are presented in Sec.~\ref{sec:appendix_data_prep}.
    \item \textbf{Ablation Study.} Additional ablation experiments analyzing the contribution of individual perceptive and temporal components are provided in Sec.~\ref{sec:appendix_ablation}.
\end{enumerate}

\section{Demo Video} 
\label{sec:demo}
In addition to the qualitative results presented in the main paper, we provide a supplementary video (\url{https://vaic-humanoid.github.io/static/videos/top.mp4}) for more detailed visualizations. The video comprehensively illustrates the efficacy of our approach, particularly highlighting robust real-world hardware deployment, real-time recording of depth-camera inputs, and the policy's generalization across various scenarios, object geometries, and weights.

\section{System Architecture and Parameters}
\label{sec:appendix_system}
\subsection{Observation Space and Depth Processing}
Our asymmetric two-stage distillation framework strictly divides the observation space. During Stage 1, the Teacher policy utilizes privileged ground-truth state information from the simulation (detailed in Table~\ref{tab:obs_space}) to master complex contact mechanics. Crucially, to ensure the teacher focuses entirely on mastering the underlying physical dynamics without being distracted by visual complexities, the depth camera is strictly disabled during this initial phase. {To maximize computational efficiency during this stage, the visual input is zero-injected, meaning the depth tensor is structurally maintained but populated entirely with zeros. Beyond preventing early visual overfitting and bypassing costly rendering overhead, this strategy guarantees dimensional consistency across the observation spaces of both stages. This structural alignment is essential, as it allows the student policy to seamlessly inherit the teacher's network architecture and pre-trained weights, ensuring a smooth and stable distillation process in Stage 2.}

The visual stream is only activated during Stage 2. Here, the Student policy replaces the dense privileged tracking inputs with a deployable observation space $o_t \in \mathcal{O}$, forcing the policy to ground its physical interactions in real-time perception. This deployable space strictly relies on:
\begin{itemize}
    \item \textbf{Proprioceptive State ($s_t^\mathrm{prop}$):} A 5-step history of joint positions, joint velocities, base angular velocity, and the projected gravity vector.
    \item \textbf{Decoupled Command ($v_t, c_t$):} A multi-axis velocity target $v_t \in \mathbb{R}^3$ and a binary interaction phase indicator $c_t \in \{0, 1\}$. To reduce the continuous observation space and mitigate policy sensitivity to minor joystick tremors, the velocity commands are quantized with a resolution of 0.2 units.
    \item \textbf{Exteroceptive Depth ($d_t$):} A temporal stream of downsampled depth images ($64 \times 36$) captured by the onboard camera at 60 Hz.
\end{itemize}

\input{tables/obs_space}
\textbf{Depth Processing and Sim-to-Real Pipeline.} To ensure robust physical deployment, we implement a rigorous depth processing pipeline that addresses sim-to-real gaps at both the sensor and signal levels. 
First, to simulate an Intel RealSense D435i camera mounted on the robot's torso (pitched downward at $48^{\circ}$), we inject extrinsic noise to the camera pose ($\sigma_{\mathrm{pos}} = 0.01$ m, $\sigma_{\mathrm{rot}} \approx 1^{\circ}$) to model installation errors, and intrinsic noise to the focal length (2\%) and aperture (5\%) via dynamic ray-casting resampling. 
Second, the raw simulated depth undergoes a sequence of signal-level augmentations: (1) \textit{Random Frame Delay} of 0 to 4 steps mimicking hardware latency; (2) \textit{Measurement Noise} via pixel-wise Gaussian perturbation ($\sigma = 0.05$ m); (3) \textit{Episodic Bias Drift} $\mathcal{U}(-0.15, 0.15)$ m accounting for thermal drift; (4) \textit{Pixel Dropout} ($p=0.1$) simulating reflective artifacts; and (5) \textit{Depth Quantization} into 100 discrete bins (equivalent to $\sim$2.9 cm accuracy) after clamping to a $[0.1, 3.0]$ m range and normalizing.

\subsection{Action Space and Low-level Control}
The policy outputs actions $a_t \in \mathbb{R}^{23}$ representing target joint position residuals $\Delta q$, such that the desired joint positions are $q_{\mathrm{target}} = q_t + \Delta q$. Note that this dimension covers the whole-body joints excluding the 6 wrist degrees of freedom (3 per hand), as they are strictly maintained or passively compliant during the macroscopic interaction tasks. These targets are dispatched to a low-level proportional-derivative (PD) controller to compute the final joint torques $\tau$:
\begin{equation}
    \tau = k_p(q_{\mathrm{target}} - q_t) - k_d \dot{q}_t
\end{equation}
where $k_p$ and $k_d$ are the joint-specific stiffness and damping gains. This residual-style position control ensures compliant and stable physical interaction.

\subsection{Network Architecture}
Our framework employs a dual-recurrent architecture to systematically decouple visual feature extraction from privileged state inference. The network consists of three core components:

\textbf{Temporal Depth Encoder:} A CNN-GRU module processes the augmented depth stream $d_t \in \mathbb{R}^{1 \times 36 \times 64}$. The spatial features are extracted via a 3-layer Convolutional Neural Network (kernel size 5, stride 2, Mish activation), flattened, and compressed via a linear layer to a 64-D spatial embedding. A subsequent Gated Recurrent Unit (GRU, hidden dimension 64) accumulates these features temporally without burn-in, outputting a 64-D temporal depth feature $\bm{z}_t^\mathrm{depth}$ followed by LayerNorm. This feature is strictly constrained for object state estimation and is not fed directly to the policy actor.

\textbf{Object Adapter:} This module explicitly predicts the interactive object's pose to decouple geometric perception from action generation. It first processes the proprioceptive observation and decoupled commands through an MLP to generate a 256-D embedding. This is concatenated with the 64-D temporal depth feature $\bm{z}_t^\mathrm{depth}$ and passed through a subsequent MLP (hidden dimensions $[256, 256]$) to output the predicted object state $\hat{\bm{s}}_t^\mathrm{obj}$. This branch is trained via supervised learning against the ground-truth object states from the simulation.

\textbf{Privilege Adapter:} To reconstruct the teacher's latent representation, this module fuses the proprioception, decoupled commands, and the geometrically transformed object point cloud feature $\hat{\bm{s}}_t^\mathrm{geo}$. The inputs are pre-processed by an MLP and fed into a core GRU module (hidden dimension 256). To prevent gradient explosion and maintain temporal stability during long-horizon interactions, a burn-in strategy is applied where gradients for the first $T/4$ steps of the sequence are detached. The output utilizes a residual skip connection (summing the GRU output and the MLP pre-processing output) before a final linear layer yields the 256-D privileged latent $\hat{\bm{z}}_t^\mathrm{priv}$ to drive the student actor.

\input{tables/reward_function}
\input{tables/ppo_hyperparam}
\section{Experimental Details}
\label{sec:appendix_exp_details}
\subsection{Training Details}
The complete reward formulation for the PPO training is detailed in Table~\ref{tab:rewards}. It is constructed to balance precise human-motion imitation with physically viable foot constraints and interaction objectives. During the student fine-tuning phase, the dense motion tracking rewards are relaxed, and the policy prioritizes task-completion objectives aligned with the decoupled velocity commands.

The networks are optimized using Proximal Policy Optimization (PPO) with Generalized Advantage Estimation (GAE). Table~\ref{tab:ppo_params} details the specific network dimensions, learning rates, and loss coefficients used during training.

\subsection{Real-World Deployment and Sim-to-Real}
We deploy the trained policies directly onto the physical humanoid robot. The inference pipeline operates onboard at 50 Hz, processing synchronized depth frames from the Intel RealSense D435i camera (captured at 60 Hz). The resulting joint position targets are dispatched to a low-level PD controller operating at 200 Hz. The specific stiffness ($k_p$) and damping ($k_d$) parameters configured for the humanoid robot hardware exactly follow the settings detailed in HAIC~\cite{li2026haic}.

To bridge the sim-to-real gap, we subject the Isaac Sim environment to extensive domain randomization during training. Table~\ref{tab:domain_rand} outlines the full distribution of randomized physical properties, including external push forces, altered mass scales, and modified joint friction, which ensure the policy remains robust to unmodeled physical perturbations during physical deployment.

\input{tables/domain_rand}

\subsection{Evaluation Metrics}
To comprehensively evaluate the interaction performance, we define two categories of metrics tracking the errors of both the humanoid root and the interactive object in the simulation environment. 

\textbf{Root State Metrics:}
\begin{itemize}
    \item \textbf{Root Position Error} ($E_{\rm rpe}$, m): The Euclidean distance between the simulated and reference root positions.
    \begin{equation}
        E_{\rm rpe} = \mathbb{E} \left[ \big\| \bm{p}^{\rm root}_t - \bm{p}^{\rm root, ref}_t \big\|_2 \right]
    \end{equation}

    \item \textbf{Root Orientation Error} ($E_{\rm roe}$, rad): The angular distance between the simulated and reference root quaternions.
    \begin{equation}
        E_{\rm roe} = \mathbb{E} \left[ 2 \arccos\left(\big| \langle \bm{q}^{\rm root}_t, \bm{q}^{\rm root, ref}_t \rangle \big|\right) \right]
    \end{equation}

    \item \textbf{Root Linear Velocity Error} ($E_{\rm rve}$, m/s): The Euclidean error of the root's linear velocity.
    \begin{equation}
        E_{\rm rve} = \mathbb{E} \left[ \big\| \bm{v}^{\rm root}_t - \bm{v}^{\rm root, ref}_t \big\|_2 \right]
    \end{equation}

    \item \textbf{Root Angular Velocity Error} ($E_{\rm rave}$, rad/s): The Euclidean error of the root's angular velocity vector.
    \begin{equation}
        E_{\rm rave} = \mathbb{E} \left[ \big\| \bm{\omega}^{\rm root}_t - \bm{\omega}^{\rm root, ref}_t \big\|_2 \right]
    \end{equation}

    \item \textbf{Root Linear Acceleration Error} ($E_{\rm rae}$, $\text{m/s}^2$): The Euclidean error of the root's linear acceleration.
    \begin{equation}
        E_{\rm rae} = \mathbb{E} \left[ \big\| \bm{a}^{\rm root}_t - \bm{a}^{\rm root, ref}_t \big\|_2 \right]
    \end{equation}
\end{itemize}

\textbf{Object State Metrics:}
\begin{itemize}
    \item \textbf{Object Position Error} ($E_{\rm ope}$, m): The Euclidean position error of the interactive object in global coordinates.
    \begin{equation}
        E_{\rm ope} = \mathbb{E} \left[ \big\| \bm{p}^{\rm obj}_t - \bm{p}^{\rm obj, ref}_t \big\|_2 \right]
    \end{equation}

    \item \textbf{Object Orientation Error} ($E_{\rm ooe}$, rad): The angular error of the object's orientation quaternion.
    \begin{equation}
        E_{\rm ooe} = \mathbb{E} \left[ 2 \arccos\left(\big| \langle \bm{q}^{\rm obj}_t, \bm{q}^{\rm obj, ref}_t \rangle \big|\right) \right]
    \end{equation}

    \item \textbf{Object Linear Velocity Error} ($E_{\rm ove}$, m/s): The Euclidean error of the object's linear velocity.
    \begin{equation}
        E_{\rm ove} = \mathbb{E} \left[ \big\| \bm{v}^{\rm obj}_t - \bm{v}^{\rm obj, ref}_t \big\|_2 \right]
    \end{equation}

    \item \textbf{Object Angular Velocity Error} ($E_{\rm oave}$, rad/s): The Euclidean error of the object's angular velocity.
    \begin{equation}
        E_{\rm oave} = \mathbb{E} \left[ \big\| \bm{\omega}^{\rm obj}_t - \bm{\omega}^{\rm obj, ref}_t \big\|_2 \right]
    \end{equation}

    \item \textbf{Object Linear Acceleration Error} ($E_{\rm oae}$, $\text{m/s}^2$): The Euclidean error of the object's linear acceleration.
    \begin{equation}
        E_{\rm oae} = \mathbb{E} \left[ \big\| \bm{a}^{\rm obj}_t - \bm{a}^{\rm obj, ref}_t \big\|_2 \right]
    \end{equation}
\end{itemize}

\section{Data Preparation and Nominal Templates}
\label{sec:appendix_data_prep}
To generate high-fidelity physical states from raw human demonstrations, we utilize an optical motion capture system followed by a rigorous retargeting pipeline. 
First, the human skeleton kinematics are mapped to the humanoid robot's 29-DoF structure using the PoseLib framework. 
Second, an Isaac Gym digital twin is established where the robot and objects are kinematically driven to extract physically valid contact labels and generate a unified $(n+m)$-body state array. 

We systematically extract unified decoupled commands directly from these reference motion trajectories. This extraction provides a dual benefit: it not only serves as the high-level task objective during training, but also establishes a highly standardized evaluation protocol. By conditioning the policy on these extracted commands during automated inference, we can record the generated kinematic trajectories and directly compute the tracking errors against the original reference motions, thereby facilitating rigorous, consistent, and reproducible benchmarking across all baselines.

Crucially for the deployable student policy, we extract a nominal point cloud template $\mathcal{P}$ for each interactive object. During Stage 2 exploration, the Privilege Adapter utilizes the implicitly inferred object state $\hat{s}_t^\mathrm{obj}$ to project this canonical template $\mathcal{P}$ into the robot's local spatial frame via $\hat{s}_t^\mathrm{geo} = \mathcal{T}(\hat{s}_t^\mathrm{obj}, \mathcal{P})$, thereby reconstructing the dense spatial priors necessary for agile interaction without relying on privileged external tracking.

\section{Ablation Study}
\label{sec:appendix_ablation}
To evaluate the contribution of individual components within the \ours{} framework, we conduct an ablation study on the high-speed Skateboard Interaction task, in which the skateboard is initialized on either side of the humanoid robot. As shown in Table~\ref{tab:ablation_model}, we progressively remove key perceptive and temporal modules to assess their impact on dynamic balance and object state estimation.

\begin{itemize}
    \item \textbf{w/o Depth:} Removing the exteroceptive depth input forces the policy to rely entirely on proprioception. This results in a catastrophic drop in success rate, as the robot cannot anticipate the initial contact side or spatial geometry of the impulsive contact.
    \item \textbf{w/o Object Adapter:} Disabling the implicit state inference module prevents the student policy from distilling the teacher's privileged geometric understanding, leading to significant divergence in object position ($E_\mathrm{ope}$) and velocity ($E_\mathrm{ove}$) tracking.
    \item \textbf{w/o GRU:} Removing the temporal recurrence degrades the policy's ability to maintain a consistent belief state of the object's latent dynamics during phase transitions, increasing root position error ($E_\mathrm{rpe}$).
\end{itemize}

\input{tables/ablation_study_on_model}

%% file: tables/obs_space.tex
\begin{table*}[ht]
\centering
\caption{Deployable Observation Space and Privileged Information for the \ours{} framework.}
\label{tab:obs_space}
\renewcommand{\arraystretch}{1.3}
\begin{tabular}{llcp{5.5cm}}
\toprule
\textbf{Category} & \textbf{Observation Term} & \textbf{Noise ($\sigma$)} & \textbf{Description} \\ 
\midrule
\multirow{10}{*}{\textbf{Student}} 
& Proprioceptive History & 0.015 & History of joint positions, velocities, and gravity vector. \\
& Base Angular Velocity & 0.05 & Base angular velocity in the local frame (steps: [0]). \\
& Previous Actions & - & Joint target history from the previous 3 steps. \\
& Decoupled Command & - & Velocity target $\bm{v}_t$ and binary interaction phase indicator $c_t$. \\
& Depth Stream & - & Sequential depth images $d_t$ processed by the CNN-GRU encoder. \\
& Obj. Geometric Prior & - & Canonical point cloud template $\mathcal{P}$ of the interactive object(s). \\
\midrule
\multirow{15}{*}{\textbf{Teacher}} 
& Clean Proprioception & 0.0 & Noise-free history of all proprioceptive states. \\
& Root \& Body Velocity & 0.0 & Linear velocity of the robot base and key bodies. \\
& Ref. Motion Tracking & 0.0 & Future reference root states, joint positions, and motion phase $\phi$. \\
& Obj. Kinematic State & 0.01 & Exact relative pose and velocity of the interactive object(s). \\
& Obj. Geometric Prior & - & Canonical point cloud template $\mathcal{P}$ of the interactive object(s). \\
& Ref. Contact Pos. & 0.01 & Exact future reference contact positions on the object. \\
& Applied Forces & - & Real applied actions and torques. \\
& Dynamics Randomization & - & Mass, friction, restitution, scale, joint armature, and damping. \\
\bottomrule
\end{tabular}

\end{table*}

%% file: tables/reward_function.tex
\begin{table}[ht]
\centering
\small
\setlength{\tabcolsep}{3pt}
\caption{Reward Functions for the \ours{} Framework. During student exploration, the policy prioritizes decoupled command tracking and phase-conditioned interaction, while dense kinematic terms are relaxed into stylistic motion priors.}
\label{tab:rewards}
\renewcommand{\arraystretch}{1.3}
\begin{tabular}{llcp{5.2cm}}
\toprule
\textbf{Term} & \textbf{Expression} & \textbf{Weight} & \textbf{Description} \\ 
\midrule
\multicolumn{4}{l}{\textit{(a) Decoupled Command Tracking}} \\
\hdashline
Linear Velocity & $\exp(-\| \bm{v}_{xy}^{\mathrm{root}} - \bm{v}_{xy}^{\mathrm{cmd}} \|^2 / \sigma_v)$ & 1.0 & Tracks the velocity command. \\
Yaw Velocity & $\exp(-\| \bm{\omega}_{z}^{\mathrm{root}} - \bm{\omega}_{z}^{\mathrm{cmd}} \|^2 / \sigma_\omega)$ & 1.0 & Tracks the yaw angular velocity command. \\

\midrule
\multicolumn{4}{l}{\textit{(b) Object Interaction Reward}} \\
\hdashline
Object Position & $\exp(-\| \bm{p}_{\mathrm{obj}} - \bm{p}_{\mathrm{obj}}^{\mathrm{ref}} \|^2 / \sigma_p)$ & 1.0 & Tracks global position of the interactive object(s). \\
Object Orientation & $\exp(-\| \bm{\theta}_{\mathrm{obj}} \ominus \bm{\theta}_{\mathrm{obj}}^{\mathrm{ref}} \|^2 / \sigma_\theta)$ & 1.0 & Tracks global orientation of the interactive object(s). \\
Phase Contact & $c_t \cdot \mathbb{E}_{o,\, e } \left[ \mathbb{I}_{o,e} \cdot r_{\mathrm{pos}}^{o,e} \cdot r_{\mathrm{force}}^{o,e} \right]$ & 1.0 & Active contact is gated by interaction indicator $c_t$. \\

\midrule
\multicolumn{4}{l}{\textit{(c) Motion Imitation Prior (Relaxed)}} \\
\hdashline
Joint Position & $\exp(-\| \bm{q} - \bm{q}^{\mathrm{ref}} \|^2 / \sigma_q)$ & 0.3 & Acts as a stylistic motion prior. \\
Joint Velocity & $\exp(-\| \dot{\bm{q}} - \dot{\bm{q}}^{\mathrm{ref}} \|^2 / \sigma_{\dot{q}})$ & 0.3 & Acts as a stylistic motion prior. \\
Upper Body Pos. & $\exp(-\| \bm{p}_{\mathrm{up}} - \bm{p}_{\mathrm{up}}^{\mathrm{ref}} \|^2 / \sigma_p)$ & 0.5 & Soft tracking for arm posture. \\
Lower Body Pos. & $\exp(-\| \bm{p}_{\mathrm{low}} - \bm{p}_{\mathrm{low}}^{\mathrm{ref}} \|^2 / \sigma_p)$ & 0.5 & Soft tracking for lower body. \\
Root Position & $\exp(-\| \bm{p}_{\mathrm{root}} - \bm{p}_{\mathrm{root}}^{\mathrm{ref}} \|^2 / \sigma_p)$ & 0.5 & Prevent root position divergence. \\
Root Orientation & $\exp(-\| \bm{\theta}_{\mathrm{root}} \ominus \bm{\theta}_{\mathrm{root}}^{\mathrm{ref}} \|^2 / \sigma_\theta)$ & 0.5 & Prevent root orientation divergence. \\

\midrule
\multicolumn{4}{l}{\textit{(d) Foot Constraints}} \\
\hdashline
Feet Air Time & $\exp( \operatorname{clip}(t_{\mathrm{air}} - t_{\mathrm{thr}}) / \sigma ) \cdot \mathbb{I}_{\mathrm{step}}$ & 0.5 & Encourages longer and more stable swing phases. \\
Feet Slip & $-\| \bm{v}_{\mathrm{foot}}^{xy} \| \cdot \mathbb{I}_{\mathrm{ground}}$ & 0.5 & Penalizes sliding velocity when feet are grounded. \\
Feet Contact & $\exp(-\| \mathbb{I}_{\mathrm{con}}^{\mathrm{real}} - \mathbb{I}_{\mathrm{con}}^{\mathrm{ref}} \|^2 / \sigma)$ & 0.5 & Matches reference rhythmic contact states. \\
Feet Air Lift & $-\sum (h_{\mathrm{foot}} < h_{\mathrm{min}}) \cdot \mathbb{I}_{\mathrm{swing}}$ & 0.5 & Penalizes tripping or dragging. \\
Impact Force & $-\| \bm{F}_{\mathrm{impact}} \|^2$ & 1.0 & Penalizes large and unsafe impact forces. \\

\midrule
\multicolumn{4}{l}{\textit{(e) Regularization}} \\
\hdashline
Action Rate & $-\| \bm{a}_t - \bm{a}_{t-1} \|^2$ & 0.1 & Penalizes rapid action changes to ensure smoothness. \\
Joint Velocity L2 & $-\| \dot{\bm{q}} \|^2$ & 5e-4 & Penalizes excessively high joint velocities (energy). \\
Joint Limits & $-\sum \operatorname{clip}(\bm{q} - \bm{q}_{\mathrm{limit}}, 0, \infty)$ & 10.0 & Penalizes exceeding physical joint limits. \\
Torque Limits & $-\sum \operatorname{clip}(\bm{\tau} - \bm{\tau}_{\mathrm{limit}}, 0, \infty)$ & 0.01 & Penalizes motor torque saturation. \\
Survival & $1.0$ & 1.0 & Constant reward for avoiding early termination. \\
\bottomrule
\end{tabular}
\vspace{-1em}
\end{table}

%% file: tables/ppo_hyperparam.tex
\begin{table}[ht]
\centering
\caption{Hyperparameters for Proximal Policy Optimization (PPO) and network architectures. The asymmetric Actor-Critic framework utilizes distinct loss coefficients to effectively balance relaxed motion imitation, privileged state distillation, and decoupled command tracking.}
\label{tab:ppo_params}
\renewcommand{\arraystretch}{1.3}
\begin{tabular}{lclc}
\toprule
\textbf{Hyperparameter} & \textbf{Value} & \textbf{Hyperparameter} & \textbf{Value} \\ 
\midrule
Optimizer & Adam & Number of Environments & 4096 \\
Rollout Steps (Horizon) & 32 & Mini-batches & 8 \\
Learning Epochs & 3 & Discount Factor ($\gamma$) & 0.99 \\
GAE Parameter ($\lambda$) & 0.95 & Clip Parameter ($\epsilon$) & 0.2 \\
Entropy Coefficient & 0.001 & Max Gradient Norm & 1.0 \\
Desired KL & 0.01 & Learning Rate & $3 \times 10^{-4}$ \\
Initial Noise Std & 1.0 & & \\
\midrule
\multicolumn{4}{l}{\textit{Loss Coefficients}} \\
\hdashline
Value Loss ($\lambda_{\mathrm{value}}$) & 1.0 & Object Loss ($\lambda_{\mathrm{obj}}$) & 1.0 \\
Privileged Loss ($\lambda_{\mathrm{priv}}$) & 1.0 & Distillation Loss ($\lambda_{\mathrm{distill}}$) & 1.0 \\
\midrule
\multicolumn{4}{l}{\textit{Network Architecture}} \\
\hdashline
Actor MLP Size & $[512, 256, 256]$ & Critic MLP Size & $[512, 256, 128]$ \\
Depth CNN Layers & 3 Conv2d & Object Adapter MLP & $[256, 256]$ \\
Depth GRU Hidden Dim & 64 & Privilege GRU Hidden Dim & 256 \\
Actor/Critic Activation & ELU & Adapter/CNN Activation & Mish \\
\bottomrule
\end{tabular}
\end{table}

%% file: tables/domain_rand.tex
\begin{table}[ht]
\centering
\caption{Domain Randomization Parameters for Robot and Objects.}
\label{tab:domain_rand}
\renewcommand{\arraystretch}{1.3}
\begin{tabular}{llc}
\toprule
\textbf{Category} & \textbf{Parameter} & \textbf{Range / Distribution} \\ 
\midrule
\multicolumn{3}{l}{\textit{\textbf{Robot Dynamics}}} \\
\hdashline
\multirow{4}{*}{Properties} 
& Link Mass Scale & $\mathcal{U}(0.9, 1.1) \times \text{default}$ \\
& Center of Mass Offset & $\mathcal{U}(-0.02, 0.02)$ m \\
& Static Friction  & $\mathcal{U}(0.3, 1.6)$ \\
& Dynamic Friction  & $\mathcal{U}(0.3, 1.2)$ \\
\midrule
\multirow{4}{*}{Actuation} 
& Joint Position Offset & $\mathcal{U}(-0.01, 0.01)$ rad \\
& Motor Stiffness Scale & $\mathcal{U}(0.9, 1.1)$ \\
& Motor Damping Scale & $\mathcal{U}(0.9, 1.1)$ \\
& Action Delay & $\mathcal{U}[40, 120]$ ms \\

\midrule
\multicolumn{3}{l}{\textit{\textbf{Object Interaction}}} \\
\hdashline
\multirow{3}{*}{{Surface}} 
& Dynamic Friction & $\mathcal{U}(0.3, 0.8)$ \\
& Static-to-Dynamic Ratio & $\mathcal{U}(1.0, 2.0)$ \\
& Restitution & $\mathcal{U}(0.0, 0.2)$ \\
\midrule
\multirow{2}{*}{{Box}} 
& Mass & $\mathcal{U}(1.0, 2.0)$ kg \\
& Scale & $\mathcal{U}(0.5, 1.2)$ \\
\midrule
\multirow{5}{*}{{Cart}} 
& Body Mass & $\mathcal{U}(11.0, 13.0)$ kg \\
& Wheel Mass & $\mathcal{U}(0.2, 0.4)$ kg \\
& Wheel Joint Friction & $\mathcal{U}(0.01, 0.1)$ N$\cdot$m \\
& Wheel Joint Damping & $\mathcal{U}(0.01, 0.1)$ N$\cdot$m$\cdot$s/rad \\
& Scale & $\mathcal{U}(0.9, 1.1)$ \\
\midrule
\multirow{5}{*}{{Skateboard}} 
& Body Mass & $\mathcal{U}(2.0, 5.0)$ kg \\
& Wheel Mass & $\mathcal{U}(0.1, 0.2)$ kg \\
& Wheel Armature & $\mathcal{U}(0.0, 1\text{e-}4)$ kg$\cdot$m$^2$ \\
& Wheel Joint Damping & $\mathcal{U}(0.0, 1\text{e-}3)$ N$\cdot$m$\cdot$s/rad \\
& Scale  & $\mathcal{U}(0.8, 1.1)$ \\
\midrule
{Slope / Stair}
& Scale  & $\mathcal{U}(0.8, 1.2)$ \\

\midrule
\multicolumn{3}{l}{\textit{\textbf{External Perturbation}}} \\
\hdashline
\multirow{2}{*}{{Push}} 
& Push Force& $\mathcal{U}(0.2, 0.5) \times \text{weight}$ \\
& Push Min Interval & 2s \\
\bottomrule
\end{tabular}
\end{table}

%% file: tables/ablation_study_on_model.tex
\begin{table}[ht]
\centering
\colorlet{lightorange}{orange!25}
\colorlet{lighterorange}{orange!12}
\caption{Quantitative ablation evaluation on \textbf{Skateboard Interaction}.}
\label{tab:ablation_model}
\resizebox{\linewidth}{!}{
\begin{tabular}{lc ccccc ccccc}
\toprule
& & \multicolumn{5}{c}{\textbf{Root State Metrics ($\downarrow$)}} & \multicolumn{5}{c}{\textbf{Object State Metrics ($\downarrow$)}} \\
\cmidrule(lr){3-7} \cmidrule(lr){8-12}
\textbf{Method} & \textbf{SR} $\uparrow$ & \textbf{$E_\mathrm{rpe}$} & \textbf{$E_\mathrm{roe}$} & \textbf{$E_\mathrm{rve}$} & \textbf{$E_\mathrm{rave}$} & \textbf{$E_\mathrm{rae}$} & \textbf{$E_\mathrm{ope}$} & \textbf{$E_\mathrm{ooe}$} & \textbf{$E_\mathrm{ove}$} & \textbf{$E_\mathrm{oave}$} & \textbf{$E_\mathrm{oae}$} \\
\midrule
w/o Depth          & 8.3\%  & 1.58$^{\pm 0.41}$ & 1.07$^{\pm 0.17}$ & 0.44$^{\pm 0.07}$ & 0.39$^{\pm 0.05}$ & 0.42$^{\pm 0.03}$ & 1.90$^{\pm 0.27}$ & 0.61$^{\pm 0.91}$ & 0.40$^{\pm 0.04}$ & 0.09$^{\pm 0.12}$ & \textbf{0.26}$^{\pm 0.03}$ \\
w/o Object Adapter & 0.0\%  & 2.08$^{\pm 0.65}$ & 1.26$^{\pm 0.19}$ & 0.50$^{\pm 0.09}$ & 0.36$^{\pm 0.02}$ & 0.39$^{\pm 0.02}$ & 1.83$^{\pm 0.14}$ & 0.31$^{\pm 0.51}$ & 0.40$^{\pm 0.03}$ & 0.07$^{\pm 0.06}$ & \textbf{0.26}$^{\pm 0.02}$ \\
w/o GRU            & 33.3\% & 1.35$^{\pm 0.66}$ & 0.95$^{\pm 0.45}$ & 0.38$^{\pm 0.13}$ & 0.35$^{\pm 0.05}$ & 0.37$^{\pm 0.09}$ & 1.46$^{\pm 0.75}$ & 0.74$^{\pm 0.85}$ & 0.36$^{\pm 0.12}$ & 0.11$^{\pm 0.09}$ & 0.27$^{\pm 0.06}$ \\
\textbf{\ours}     & \cellcolor{lightorange}\textbf{83.3\%} & \cellcolor{lightorange}\textbf{0.62$^{\pm 0.12}$} & \cellcolor{lightorange}\textbf{0.38$^{\pm 0.19}$} & \cellcolor{lightorange}\textbf{0.29$^{\pm 0.05}$} & \cellcolor{lightorange}\textbf{0.27$^{\pm 0.07}$} & \cellcolor{lightorange}\textbf{0.35$^{\pm 0.07}$} & \cellcolor{lightorange}\textbf{0.62$^{\pm 0.17}$} & \cellcolor{lightorange}\textbf{0.07$^{\pm 0.06}$} & \cellcolor{lightorange}\textbf{0.31$^{\pm 0.10}$} & \cellcolor{lightorange}\textbf{0.02$^{\pm 0.02}$} & \cellcolor{lighterorange}0.27$^{\pm 0.04}$ \\
\bottomrule
\end{tabular}}
\end{table}

%% file: example.bib
@article{long2024learning,
  title={Learning Humanoid Locomotion with Perceptive Internal Model},
  author={Long, Junfeng and Ren, Junli and Shi, Moji and Wang, Zirui and Huang, Tao and Luo, Ping and Pang, Jiangmiao},
  journal={arXiv preprint arXiv:2411.14386},
  year={2024}
}

@inproceedings{zhuang2024humanoid,
  title={Humanoid Parkour Learning},
  author={Zhuang, Ziwen and Yao, Shenzhe and Zhao, Hang},
  booktitle={Conference on Robot Learning (CoRL)},
  year={2024}
}

@inproceedings{he2024hover,
  title={Hover: Versatile neural whole-body controller for humanoid robots},
  author={He, Tairan and Xiao, Wenli and Lin, Toru and Luo, Zhengyi and Xu, Zhenjia and Jiang, Zhenyu and Kautz, Jan and Liu, Changliu and Shi, Guanya and Wang, Xiaolong and others},
  booktitle={International Conference on Robotics and Automation (ICRA)},
  year={2025}
}

@inproceedings{dao2024sim,
  title={Sim-to-real learning for humanoid box loco-manipulation},
  author={Dao, Jeremy and Duan, Helei and Fern, Alan},
  booktitle={International Conference on Robotics and Automation (ICRA)},
  year={2024},
}

@article{schulman2017proximalpolicyoptimizationalgorithms,
  title={Proximal policy optimization algorithms},
  author={Schulman, John and Wolski, Filip and Dhariwal, Prafulla and Radford, Alec and Klimov, Oleg},
  journal={arXiv preprint arXiv:1707.06347},
  year={2017}
}

@inproceedings{bhatnagar2022behave,
    title     = {{BEHAVE}: Dataset and Method for Tracking Human Object Interactions},
    author    = {Bhatnagar, Bharat Lal and Xie, Xianghui and Petrov, Ilya and Sminchisescu, Cristian and Theobalt, Christian and Pons-Moll, Gerard},
    booktitle = {Proceedings of the Computer Vision and Pattern Recognition Conference (CVPR)},
    year      = {2022},
}

@inproceedings{xu2025interact,
  title = {InterAct: Advancing Large-Scale Versatile 3D Human-Object Interaction Generation},
  author = {Xu, Sirui and Li, Dongting and Zhang, Yucheng and Xu, Xiyan and Long, Qi and Wang, Ziyin and Lu, Yunzhi and Dong, Shuchang and Jiang, Hezi and Gupta, Akshat and Wang, Yu-Xiong and Gui, Liang-Yan},
  booktitle = {Proceedings of the Computer Vision and Pattern Recognition Conference (CVPR)},
  year = {2025},
}

@article{wang2023physhoi,
  author    = {Wang, Yinhuai and Lin, Jing and Zeng, Ailing and Luo, Zhengyi and Zhang, Jian and Zhang, Lei},
  title     = {PhysHOI: Physics-Based Imitation of Dynamic Human-Object Interaction},
  journal   = {arXiv preprint arXiv:2312.04393},
  year      = {2023},
}

@inproceedings{pan2025tokenhsi,
  title={TokenHSI: Unified Synthesis of Physical Human-Scene Interactions through Task Tokenization},
  author={Pan, Liang and Yang, Zeshi and Dou, Zhiyang and Wang, Wenjia and Huang, Buzhen and Dai, Bo and Komura, Taku and Wang, Jingbo},
  booktitle={Proceedings of the Computer Vision and Pattern Recognition Conference (CVPR)},
  year={2025},
}

@article{joao2025gmr,
  title={Retargeting Matters: General Motion Retargeting for Humanoid Motion Tracking},
  author= {Joao Pedro Araujo and Yanjie Ze and Pei Xu and Jiajun Wu and C. Karen Liu},
  year= {2025},
  journal= {arXiv preprint arXiv:2510.02252}
}

@article{he2025asap,
  title={ASAP: Aligning Simulation and Real-World Physics for Learning Agile Humanoid Whole-Body Skills},
  author={He, Tairan and Gao, Jiawei and Xiao, Wenli and Zhang, Yuanhang and Wang, Zi and Wang, Jiashun and Luo, Zhengyi and He, Guanqi and Sobanbabu, Nikhil and Pan, Chaoyi and Yi, Zeji and Qu, Guannan and Kitani, Kris and Hodgins, Jessica and Fan, Linxi "Jim" and Zhu, Yuke and Liu, Changliu and Shi, Guanya},
  journal={arXiv preprint arXiv:2502.01143},
  year={2025}
}

@article{liao2025beyondmimicmotiontrackingversatile,
  title={Beyondmimic: From motion tracking to versatile humanoid control via guided diffusion},
  author={Liao, Qiayuan and Truong, Takara E and Huang, Xiaoyu and Gao, Yuman and Tevet, Guy and Sreenath, Koushil and Liu, C Karen},
  journal={arXiv preprint arXiv:2508.08241},
  year={2025}
}

@inproceedings{allshire2025videomimic,
  title     = {Visual imitation enables contextual humanoid control},
  author    = {Allshire, Arthur and Choi, Hongsuk and Zhang, Junyi and McAllister, David and Zhang, Anthony and Kim, Chung Min and Darrell, Trevor and Abbeel, Pieter and Malik, Jitendra and Kanazawa, Angjoo},
  booktitle = {Proceedings of the Conference on Robot Learning (CoRL)},
  year      = {2025}
}

@inproceedings{xu2025intermimic,
      title={InterMimic: Towards Universal Whole-Body Control for Physics-Based Human-Object Interactions},
      author={Xu, Sirui and Ling, Hung Yu and Wang, Yu-Xiong and Gui, Liangyan},
      booktitle={Proceedings of the Computer Vision and Pattern Recognition Conference (CVPR)},
      year={2025}
}

@article{shao2025visualmimic,
title={VisualMimic: Visual Humanoid Loco-Manipulation via Motion Tracking and Generation},
author= {Shaofeng Yin and Yanjie Ze and Hong-Xing Yu and C. Karen Liu and Jiajun Wu},
year= {2025},
journal= {arXiv preprint arXiv:2509.20322}
}

@article{weng2025hdmilearninginteractivehumanoid,
  title={Hdmi: Learning interactive humanoid whole-body control from human videos},
  author={Weng, Haoyang and Li, Yitang and Sobanbabu, Nikhil and Wang, Zihan and Luo, Zhengyi and He, Tairan and Ramanan, Deva and Shi, Guanya},
  journal={arXiv preprint arXiv:2509.16757},
  year={2025}
}

@article{zhao2025resmimicgeneralmotiontracking,
  title={Resmimic: From general motion tracking to humanoid whole-body loco-manipulation via residual learning},
  author={Zhao, Siheng and Ze, Yanjie and Wang, Yue and Liu, C Karen and Abbeel, Pieter and Shi, Guanya and Duan, Rocky},
  journal={arXiv preprint arXiv:2510.05070},
  year={2025}
}

@article{zhang2025falcon,
          title={FALCON: Learning Force-Adaptive Humanoid Loco-Manipulation},
          author={Zhang, Yuanhang and Yuan, Yifu and Gurunath, Prajwal and Gupta, Ishita and Omidshafiei, Shayegan and Agha-mohammadi, Ali-akbar and Vazquez-Chanlatte, Marcell and Pedersen, Liam and He, Tairan and Shi, Guanya},
          journal={arXiv preprint arXiv:2505.06776},
          year={2025}
        }

@article{he2025viral,
  title={VIRAL: Visual Sim-to-Real at Scale for Humanoid Loco-Manipulation},
  author={He, Tairan and Wang, Zi and Xue, Haoru and Ben, Qingwei and Luo, Zhengyi and Xiao, Wenli and Yuan, Ye and Da, Xingye and Casta{\~n}eda, Fernando and Sastry, Shankar and others},
  journal={arXiv preprint arXiv:2511.15200},
  year={2025}
}

@article{jiang2025wholebodyvlaunifiedlatentvla,
  title={WholeBodyVLA: Towards Unified Latent VLA for Whole-Body Loco-Manipulation Control},
  author={Jiang, Haoran and Chen, Jin and Bu, Qingwen and Chen, Li and Shi, Modi and Zhang, Yanjie and Li, Delong and Suo, Chuanzhe and Wang, Chuang and Peng, Zhihui and others},
  journal={arXiv preprint arXiv:2512.11047},
  year={2025}
}

@article{luo2025sonic,
  title={Sonic: Supersizing motion tracking for natural humanoid whole-body control},
  author={Luo, Zhengyi and Yuan, Ye and Wang, Tingwu and Li, Chenran and Chen, Sirui and Casta{\~n}eda, Fernando and Cao, Zi-Ang and Li, Jiefeng and Minor, David and Ben, Qingwei and others},
  journal={arXiv preprint arXiv:2511.07820},
  year={2025}
}

@article{zeng2025behavior,
  title={Behavior foundation model for humanoid robots},
  author={Zeng, Weishuai and Lu, Shunlin and Yin, Kangning and Niu, Xiaojie and Dai, Minyue and Wang, Jingbo and Pang, Jiangmiao},
  journal={arXiv preprint arXiv:2509.13780},
  year={2025}
}

@article{yuan2025behavior,
  title={Behavior Foundation Model: Towards Next-Generation Whole-Body Control System of Humanoid Robots},
  author={Yuan, Mingqi and Yu, Tao and Ge, Wenqi and Yao, Xiuyong and Li, Dapeng and Wang, Huijiang and Chen, Jiayu and Jin, Xin and Li, Bo and Chen, Hua and others},
  journal={arXiv preprint arXiv:2506.20487},
  year={2025}
}

@article{li2025bfm,
  title={BFM-Zero: A Promptable Behavioral Foundation Model for Humanoid Control Using Unsupervised Reinforcement Learning},
  author={Li, Yitang and Luo, Zhengyi and Zhang, Tonghe and Dai, Cunxi and Kanervisto, Anssi and Tirinzoni, Andrea and Weng, Haoyang and Kitani, Kris and Guzek, Mateusz and Touati, Ahmed and others},
  journal={arXiv preprint arXiv:2511.04131},
  year={2025}
}

@article{jiang2025uniact,
  title={UniAct: Unified Motion Generation and Action Streaming for Humanoid Robots},
  author={Jiang, Nan and He, Zimo and Yu, Wanhe and Pang, Lexi and Li, Yunhao and Li, Hongjie and Cui, Jieming and Li, Yuhan and Wang, Yizhou and Zhu, Yixin and others},
  journal={arXiv preprint arXiv:2512.24321},
  year={2025}
}

@article{xue2025leverb,
  title={Leverb: Humanoid whole-body control with latent vision-language instruction},
  author={Xue, Haoru and Huang, Xiaoyu and Niu, Dantong and Liao, Qiayuan and Kragerud, Thomas and Gravdahl, Jan Tommy and Peng, Xue Bin and Shi, Guanya and Darrell, Trevor and Sreenath, Koushil and others},
  journal={arXiv preprint arXiv:2506.13751},
  year={2025}
}

@article{yang2025omniretarget,
  title={Omniretarget: Interaction-preserving data generation for humanoid whole-body loco-manipulation and scene interaction},
  author={Yang, Lujie and Huang, Xiaoyu and Wu, Zhen and Kanazawa, Angjoo and Abbeel, Pieter and Sferrazza, Carmelo and Liu, C Karen and Duan, Rocky and Shi, Guanya},
  journal={arXiv preprint arXiv:2509.26633},
  year={2025}
}

@article{wu2024thor,
  title={Thor: Text to human-object interaction diffusion via relation intervention},
  author={Wu, Qianyang and Shi, Ye and Huang, Xiaoshui and Yu, Jingyi and Xu, Lan and Wang, Jingya},
  journal={arXiv preprint arXiv:2403.11208},
  year={2024}
}

@article{chen2025gmt,
  title={GMT: General Motion Tracking for Humanoid Whole-Body Control},
  author={Chen, Zixuan and Ji, Mazeyu and Cheng, Xuxin and Peng, Xuanbin and Peng, Xue Bin and Wang, Xiaolong},
  journal={arXiv preprint arXiv:2506.14770},
  year={2025}
}

@article{yin2025unitrackerlearninguniversalwholebody,
  title={Unitracker: Learning universal whole-body motion tracker for humanoid robots},
  author={Yin, Kangning and Zeng, Weishuai and Fan, Ke and Dai, Minyue and Wang, Zirui and Zhang, Qiang and Tian, Zheng and Wang, Jingbo and Pang, Jiangmiao and Zhang, Weinan},
  journal={arXiv preprint arXiv:2507.07356},
  year={2025}
}

@article{sun2025ulc,
  title={Ulc: A unified and fine-grained controller for humanoid loco-manipulation},
  author={Sun, Wandong and Feng, Luying and Cao, Baoshi and Liu, Yang and Jin, Yaochu and Xie, Zongwu},
  journal={arXiv preprint arXiv:2507.06905},
  year={2025}
}

@article{wei2025hmc,
  title={HMC: Learning Heterogeneous Meta-Control for Contact-Rich Loco-Manipulation},
  author={Wei, Lai and Peng, Xuanbin and Qiu, Ri-Zhao and Huang, Tianshu and Cheng, Xuxin and Wang, Xiaolong},
  journal={arXiv preprint arXiv:2511.14756},
  year={2025}
}

@article{peng2018deepmimic,
  title={Deepmimic: Example-guided deep reinforcement learning of physics-based character skills},
  author={Peng, Xue Bin and Abbeel, Pieter and Levine, Sergey and Van de Panne, Michiel},
  journal={ACM Transactions On Graphics (TOG)},
  volume={37},
  number={4},
  pages={1--14},
  year={2018},
  publisher={ACM New York, NY, USA}
}

@article{peng2021amp,
  title={Amp: Adversarial motion priors for stylized physics-based character control},
  author={Peng, Xue Bin and Ma, Ze and Abbeel, Pieter and Levine, Sergey and Kanazawa, Angjoo},
  journal={ACM Transactions on Graphics (ToG)},
  volume={40},
  number={4},
  pages={1--20},
  year={2021},
  publisher={ACM New York, NY, USA}
}

@article{shao2025langwbc,
  title={LangWBC: Language-directed Humanoid Whole-Body Control via End-to-end Learning},
  author={Shao, Yiyang and Huang, Xiaoyu and Zhang, Bike and Liao, Qiayuan and Gao, Yuman and Chi, Yufeng and Li, Zhongyu and Shao, Sophia and Sreenath, Koushil},
  journal={arXiv preprint arXiv:2504.21738},
  year={2025}
}

@article{lin2025simgenhoiphysicallyrealisticwholebody,
      title={SimGenHOI: Physically Realistic Whole-Body Humanoid-Object Interaction via Generative Modeling and Reinforcement Learning}, 
      author={Yuhang Lin and Yijia Xie and Jiahong Xie and Yuehao Huang and Ruoyu Wang and Jiajun Lv and Yukai Ma and Xingxing Zuo},
      year={2025},
      journal={arXiv preprint arXiv:2508.14120},
}

@article{li2026haic,
  title = {HAIC: Humanoid Agile Object Interaction Control via Dynamics-Aware World Model},
  author = {Li, Dongting and Chen, Xingyu and Wu, Qianyang and Chen, Bo and Wu, Sikai and Wu, Hanyu and Zhang, Guoyao and Li, Liang and Zhou, Mingliang and Xiang, Diyun and Ma, Jianzhu and Zhang, Qiang and Xu, Renjing},
  journal = {arXiv preprint arXiv:2602.11758},
  year = {2026}
}

@article{sun2025dpl,
  title={DPL: Depth-only Perceptive Humanoid Locomotion via Realistic Depth Synthesis and Cross-Attention Terrain Reconstruction},
  author={Jingkai Sun and Gang Han and Pihai Sun and Wen Zhao and Jiahang Cao and Jiaxu Wang and Yijie Guo and Qiang Zhang},
  journal={arXiv preprint arXiv:2510.07152}, 
  year={2025}
}

@article{han2025prior,
  title={PRIOR: Perceptive Learning for Humanoid Locomotion with Reference Gait Priors},
  author={Chenxi Han and Shilu He and Yi Cheng and Linqi Ye and Houde Liu},
  journal={arXiv preprint arXiv:2603.18979},
  year={2026}
}

@inproceedings{agarwal2023legged,
  title={Legged Locomotion in Challenging Terrains using Egocentric Vision},
  author={Ananye Agarwal and Ashish Kumar and Jitendra Malik and Deepak Pathak},
  booktitle={Conference on Robot Learning (CoRL)},
  year={2022}
}

@inproceedings{liu2025ego,
  title={Ego-Vision World Model for Humanoid Contact Planning},
  author={Hang Liu and Yuman Gao and Sangli Teng and Yufeng Chi and Yakun Sophia Shao and Zhongyu Li and Maani Ghaffari and Koushil Sreenath},
  booktitle={International Conference on Robotics and Automation (ICRA)}, 
  year={2026}
}

@article{wu2026sugar,
  title={SUGAR: A Scalable Human-Video-Driven Generalizable Humanoid Loco-Manipulation Learning Framework},
  author={Wu, Tianshu and Kong, Xiangqi and Chen, Yue and Yu, Qize and Ye, Hang and Li, Jia and Wang, Yizhou and Dong, Hao},
  journal={arXiv preprint arXiv:2605.20373},
  year={2026}
}

@article{zhang2026rpl,
  title={RPL: Learning Robust Humanoid Perceptive Locomotion on Challenging Terrains},
  author={Zhang, Yuanhang and Seo, Younggyo and Chen, Juyue and Yuan, Yifu and Sreenath, Koushil and Abbeel, Pieter and Sferrazza, Carmelo and Liu, Karen and Duan, Rocky and Shi, Guanya},
  journal={arXiv preprint arXiv:2602.03002},
  year={2026}
}

@article{wu2026perceptive,
  title={Perceptive humanoid parkour: Chaining dynamic human skills via motion matching},
  author={Wu, Zhen and Huang, Xiaoyu and Yang, Lujie and Zhang, Yuanhang and Sreenath, Koushil and Chen, Xi and Abbeel, Pieter and Duan, Rocky and Kanazawa, Angjoo and Sferrazza, Carmelo and others},
  journal={arXiv preprint arXiv:2602.15827},
  year={2026}
}

@article{zhuang2026deep,
  title={Deep Whole-body Parkour},
  author={Zhuang, Ziwen and Zhu, Shaoting and Zhao, Mengjie and Zhao, Hang},
  journal={arXiv preprint arXiv:2601.07701},
  year={2026}
}

@article{zhu2026ttt,
  title={TTT-Parkour: Rapid Test-Time Training for Perceptive Robot Parkour},
  author={Zhu, Shaoting and Ye, Baijun and Wang, Jiaxuan and Chen, Jiakang and Zhuang, Ziwen and Mou, Linzhan and Huang, Runhan and Zhao, Hang},
  journal={arXiv preprint arXiv:2602.02331},
  year={2026}
}

@article{chen2026learning,
  title={Learning human-like badminton skills for humanoid robots},
  author={Chen, Yeke and Dong, Shihao and Ji, Xiaoyu and Sun, Jingkai and Luo, Zeren and Zhao, Liu and Zhang, Jiahui and Li, Wanyue and Ma, Ji and Xu, Bowen and others},
  journal={arXiv preprint arXiv:2602.08370},
  year={2026}
}

@article{zhang2026learning,
  title={Learning athletic humanoid tennis skills from imperfect human motion data},
  author={Zhang, Zhikai and Lu, Haofei and Lian, Yunrui and Chen, Ziqing and Liu, Yun and Lin, Chenghuai and Xue, Han and Zeng, Zicheng and Qi, Zekun and Zheng, Shaolin and others},
  journal={arXiv preprint arXiv:2603.12686},
  year={2026}
}

@article{kong2026learning,
  title={Learning Soccer Skills for Humanoid Robots: A Progressive Perception-Action Framework},
  author={Kong, Jipeng and Liu, Xinzhe and Lin, Yuhang and Han, Jinrui and Schwertfeger, S{\"o}ren and Bai, Chenjia and Li, Xuelong},
  journal={arXiv preprint arXiv:2602.05310},
  year={2026}
}

@article{ren2026smash,
  title={SMASH: Mastering Scalable Whole-Body Skills for Humanoid Ping-Pong with Egocentric Vision},
  author={Ren, Junli and Li, Yinghui and Zhang, Kai and Fu, Penglin and Jiang, Haoran and Pan, Yixuan and Zeng, Guangjun and Huang, Tao and Guo, Weizhong and Lu, Peng and others},
  journal={arXiv preprint arXiv:2604.01158},
  year={2026}
}

@article{he2026ultra,
  title={ULTRA: Unified Multimodal Control for Autonomous Humanoid Whole-Body Loco-Manipulation},
  author={He, Xialin and Xu, Sirui and Li, Xinyao and Dong, Runpei and Bian, Liuyu and Wang, Yu-Xiong and Gui, Liang-Yan},
  journal={arXiv preprint arXiv:2603.03279},
  year={2026}
}

@article{lin2026pro,
  title={Pro-HOI: Perceptive Root-guided Humanoid-Object Interaction},
  author={Lin, Yuhang and Shi, Jiyuan and Wang, Dewei and Kong, Jipeng and Liu, Yong and Bai, Chenjia and Li, Xuelong},
  journal={arXiv preprint arXiv:2603.01126},
  year={2026}
}

@article{lin2026lessmimic,
  title={Lessmimic: Long-horizon humanoid interaction with unified distance field representations},
  author={Lin, Yutang and Cui, Jieming and Li, Yixuan and Jia, Baoxiong and Zhu, Yixin and Huang, Siyuan},
  journal={arXiv preprint arXiv:2602.21723},
  year={2026}
}

@article{sun2026now,
  title={Now You See That: Learning End-to-End Humanoid Locomotion from Raw Pixels},
  author={Sun, Wandong and Su, Yongbo and Huang, Leoric and Zhang, Alex and Wei, Dwyane and San, Mu and Tian, Daniel and Cao, Ellie and Cao, Baoshi and Liu, Yang and others},
  journal={arXiv preprint arXiv:2602.06382},
  year={2026}
}

@article{rempe2026kimodo,
  title={Kimodo: Scaling controllable human motion generation},
  author={Rempe, Davis and Petrovich, Mathis and Yuan, Ye and Zhang, Haotian and Peng, Xue Bin and Jiang, Yifeng and Wang, Tingwu and Iqbal, Umar and Minor, David and de Ruyter, Michael and others},
  journal={arXiv preprint arXiv:2603.15546},
  year={2026}
}

@article{wang2026humanx,
  title={HumanX: Toward Agile and Generalizable Humanoid Interaction Skills from Human Videos},
  author={Wang, Yinhuai and Zhao, Qihan and Lau, Yuen Fui and Yu, Runyi and Tsui, Hok Wai and Chen, Qifeng and Wang, Jingbo and Pang, Jiangmiao and Tan, Ping},
  journal={arXiv preprint arXiv:2602.02473},
  year={2026}
}

@inproceedings{zhang2023neuraldome,
  title={Neuraldome: A neural modeling pipeline on multi-view human-object interactions},
  author={Zhang, Juze and Luo, Haimin and Yang, Hongdi and Xu, Xinru and Wu, Qianyang and Shi, Ye and Yu, Jingyi and Xu, Lan and Wang, Jingya},
  booktitle={Proceedings of the IEEE/CVF Conference on Computer Vision and Pattern Recognition},
  pages={8834--8845},
  year={2023}
}

@inproceedings{zhao2024m,
  title={I'm hoi: Inertia-aware monocular capture of 3d human-object interactions},
  author={Zhao, Chengfeng and Zhang, Juze and Du, Jiashen and Shan, Ziwei and Wang, Junye and Yu, Jingyi and Wang, Jingya and Xu, Lan},
  booktitle={Proceedings of the IEEE/CVF Conference on Computer Vision and Pattern Recognition},
  pages={729--741},
  year={2024}
}

@inproceedings{lu2025humoto,
  title={HUMOTO: A 4d dataset of mocap human object interactions},
  author={Lu, Jiaxin and Huang, Chun-Hao Paul and Bhattacharya, Uttaran and Huang, Qixing and Zhou, Yi},
  booktitle={Proceedings of the IEEE/CVF International Conference on Computer Vision},
  pages={10886--10897},
  year={2025}
}

@article{wang2025physhsi,
  title={Physhsi: Towards a real-world generalizable and natural humanoid-scene interaction system},
  author={Wang, Huayi and Zhang, Wentao and Yu, Runyi and Huang, Tao and Ren, Junli and Jia, Feiyu and Wang, Zirui and Niu, Xiaojie and Chen, Xiao and Chen, Jiahe and others},
  journal={arXiv preprint arXiv:2510.11072},
  year={2025}
}
